\documentclass[final]{cvpr}

\usepackage{times}
\usepackage{epsfig}
\usepackage{graphicx}
\usepackage{booktabs}
\usepackage{multirow}
\usepackage{amsmath}
\usepackage{amssymb}
\usepackage{setspace}
\usepackage{bm}
\usepackage{subfigure}
\usepackage{algorithmic, algorithm}
\usepackage[pagebackref=true,breaklinks=true,colorlinks,bookmarks=false]{hyperref}

\renewcommand{\paragraph}{\textbf}
\expandafter\def\expandafter\normalsize\expandafter{%
\normalsize\setlength\abovedisplayskip{3pt}}
\expandafter\def\expandafter\normalsize\expandafter{%
\normalsize\setlength\belowdisplayskip{3pt}}

\def\cvprPaperID{5248} 
\def\confYear{CVPR 2021}

\begin{document}

\title{Prototype-supervised Adversarial Network for Targeted Attack of Deep Hashing}

\author{Xunguang Wang\textsuperscript{1,$\ast$},
	Zheng Zhang\textsuperscript{1,2,$\ast$,$\dagger$}, Baoyuan Wu\textsuperscript{3,4}, Fumin Shen\textsuperscript{5,6}, Guangming Lu\textsuperscript{1} \\
	\textsuperscript{1}Harbin Institute of Technology, Shenzhen, \textsuperscript{2}Peng Cheng Laboratory \\ \textsuperscript{3}School of Data Science, The Chinese University of Hong Kong, Shenzhen \\ \textsuperscript{4}Secure Computing Lab of Big Data, Shenzhen Research Institute of Big Data \\ \textsuperscript{5}University of Electronic Science and Technology of China, \textsuperscript{6}Koala Uran Tech. \\
	{\tt\small \{xunguangwang, darrenzz219, wubaoyuan1987, fumin.shen\}@gmail.com, luguangm@hit.edu.cn}
}


\maketitle
\renewcommand{\thefootnote}{\fnsymbol{footnote}}
\footnotetext[1]{Equal contribution}
\footnotetext[2]{Corresponding author}

\pagestyle{empty}  
\thispagestyle{empty} 

\begin{abstract}
	Due to its powerful capability of representation learning and high-efficiency computation, deep hashing has made significant progress in large-scale image retrieval. However, deep hashing networks are vulnerable to adversarial examples, which is a practical secure problem but seldom studied in hashing-based retrieval field. In this paper, we propose a novel prototype-supervised adversarial network (ProS-GAN), which formulates a flexible generative architecture for efficient and effective targeted hashing attack. To the best of our knowledge, this is the first generation-based method to attack deep hashing networks. Generally, our proposed framework consists of three parts, i.e., a PrototypeNet, a generator and a discriminator. Specifically, the designed PrototypeNet embeds the target label into the semantic representation and learns the prototype code as the category-level representative of the target label. Moreover, the semantic representation and the original image are jointly fed into the generator for flexible targeted attack. Particularly, the prototype code is adopted to supervise the generator to construct the targeted adversarial example by minimizing the Hamming distance between the hash code of the adversarial example and the prototype code. Furthermore, the generator is against the discriminator to simultaneously encourage the adversarial examples visually realistic and the semantic representation informative. Extensive experiments verify that the proposed framework can efficiently produce adversarial examples with better targeted attack performance and transferability over state-of-the-art targeted attack methods of deep hashing.
\end{abstract}

\begin{figure*}
	\begin{center}
		\includegraphics[width=0.9\linewidth]{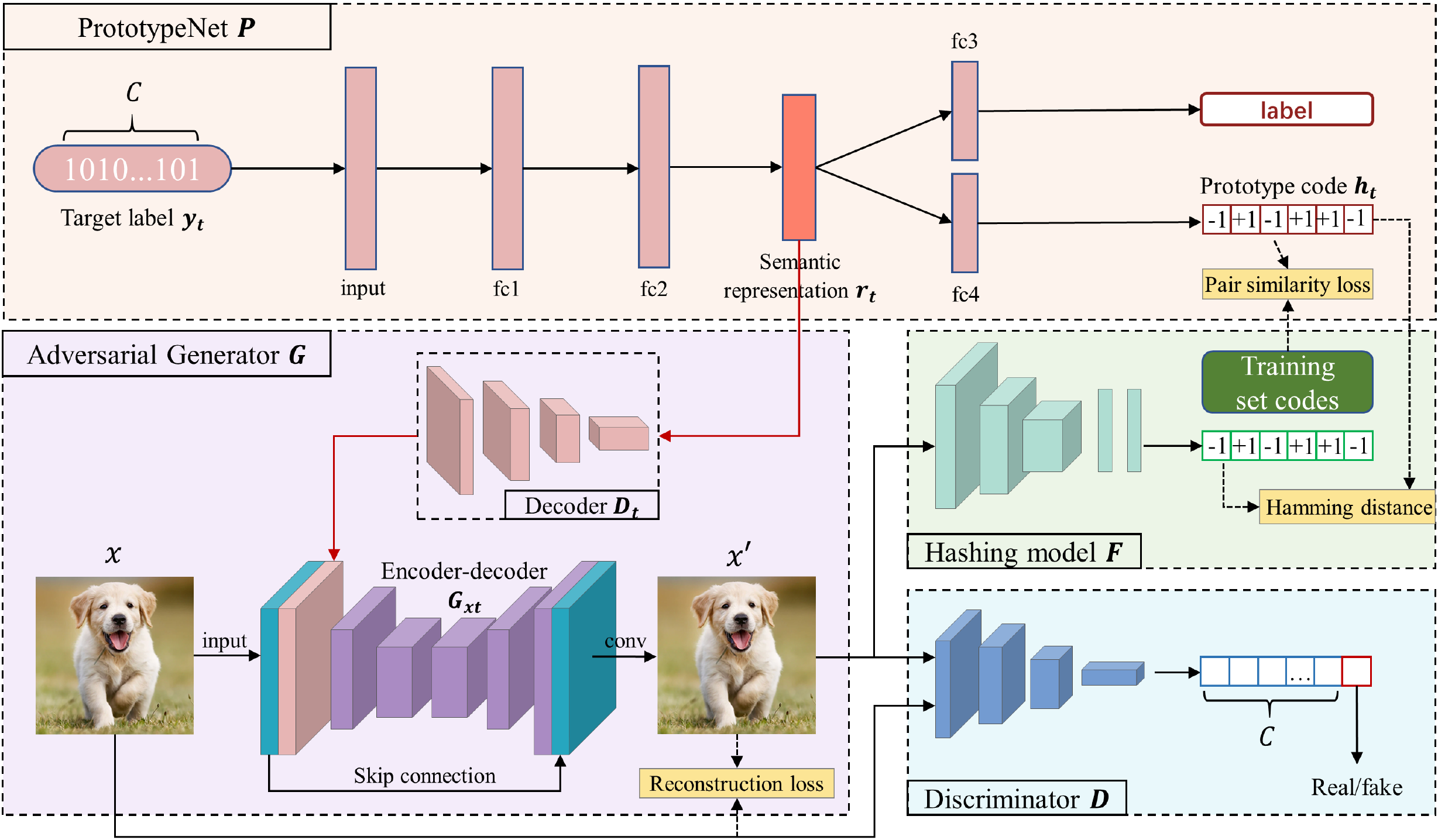}
	\end{center}
	\vspace{-0.15cm}
	\caption{The framework of our Prototype-supervised Adversarial Network (ProS-GAN).}
	\label{fig:framework}
	\vspace{-0.2cm}
\end{figure*}

\vspace{-0.2cm}
\section{Introduction}
With the explosive growth of high-dimensional and large-scale multimedia data, approximate nearest neighbor (ANN) search \cite{andoni2006near} has attracted much attention in information retrieval due to its efficiency and effectiveness.
As a solution of ANN, hashing \cite{wang2017survey} maps high-dimension data to compact binary codes meanwhile preserving the semantic similarities, yielding significant advantages in storage cost and retrieval speed.
Benefiting from the strong representation ability of deep learning, deep hashing that employs deep neural networks (DNNs) to automatically extract features has achieved great success in learning to hash \cite{xia2014supervised, lai2015simultaneous, zhu2016deep, li2016feature, liu2016deep, cao2017hashnet}, and also has been demonstrated its superior performance than the shallow hashing methods.

Notably, recent studies \cite{szegedy2013intriguing, goodfellow2014explaining, kurakin2016adversarial, madry2017towards, carlini2017towards} have recognized that DNNs are usually vulnerable to adversarial examples, which are intentionally perturbed by adding imperceptible noises to original images but can fool the networks to make incorrect predictions.
Deep hashing methods have achieved encouraging performance on many benchmarks, while, at the same time, they inevitably inherit the fundamental fragility of DNNs on handling adversarial examples \cite{yang2018adversarial, bai2020targeted}. This imperceptible malicious attack poses a serious security threat to the deep hashing-based image retrieval.
For example, when querying with an intentionally perturbed dog image, a hashing based retrieval system may return violent images.
Accordingly, it is necessary to study the adversarial attacks on deep hashing models in order to recognize their flaws and help solve their security risks.

Currently, many works about adversarial examples have been studied in image classification, but \textit{very few} researches focus on the security of deep hashing based retrieval.
Different from the typical classification, hashing aims to learn the semantic similarity between images, and its final outputs are discrete binary codes instead of categories.
Thus, these attack methods in image classification cannot be directly used or transferred to the deep hashing tasks.
Existing adversarial attack methods of deep hashing only include an non-targeted attack method called HAG \cite{yang2018adversarial} and two targeted attack methods called P2P \cite{bai2020targeted} and DHTA \cite{bai2020targeted}, respectively.
Notwithstanding, they are verified to be effective in attack, there are still some significant limitations hindering the current adversarial attacks in deep hashing.
On one hand, these methods are inefficient because they are optimization-based methods that rely on a very time-consuming iterative gradient.
For example, to make the attacked hashing model bias significantly, DHTA requires around 2000 iterations to optimize adversarial perturbations.
On the other hand, these methods heuristically select a hash code as representative of the target label to guide the generation of the targeted adversarial example.
However, this code can not represent the discriminative category-level semantics of the target label due to lack of preserving similarity with relevant labels and dissimilarity with irrelevant labels.
Therefore, how to construct more representative semantic-preserving hash code of the target label becomes crucially important to achieve satisfactory performance in adversarial attack of deep hashing.

To overcome the above deficiencies and inspired by generation-based adversarial attacks \cite{baluja2017adversarial, xiao2018generating, han2019once} in classification, this paper proposes a prototype-supervised adversarial network (ProS-GAN) for efficient and effective targeted attack in deep hashing based retrieval.
By feeding an original image and a target label into ProS-GAN, it can generate the targeted adversarial example, which would mislead the attacked hashing network to retrieve the images semantically related to the target label.
Specifically, ProS-GAN is composed of three sub-networks: a prototype network (PrototypeNet), an adversarial generator network and a discriminator network, as shown in Figure \ref{fig:framework}.
The designed PrototypeNet encodes the input target label into the semantic representation and learns the prototype code as the representative of the target label. 
Then, the adversarial generator incorporates the original image and the semantic representation into a self-reconstruction network for generating the adversarial example.
Moreover, the prototype code is adopted to supervise the generator to construct the targeted adversarial example by minimizing the Hamming distance between the hash code of the adversarial example and the prototype code.
In addition, the discriminator is used to distinguish real/fake images and categorize them into the original and target categories, respectively.
The generator and discriminator are trained in an adversarial manner to encourage the generated images visually realistic and the semantic representation informative for further improving the targeted attack performance.
In summary, the main contributions are outlined as follows:
\begin{itemize}
	\item We propose a novel prototype-supervised adversarial network\footnote{Code: \url{https://github.com/xunguangwang/ProS-GAN}} (ProS-GAN) for flexible targeted hashing attack. \textit{To the best of our knowledge, this is the very first work of formulating a generative architecture for arbitrary-target attack in deep hashing based retrieval.} Importantly, different from the existing attack methods, our work could efficiently and effectively generate preferable adversarial examples with robust transferability under one-forward pass.
	
	\item Instead of heuristically selecting a hash code as representative of the target label, we leverage the invariant side of semantics to generate the flexible prototype code in our ProtypeNet as the expected mainstay of the target label from the optimization view.
	
	\item Extensive experiments validate the superior efficiency and transferability of the produced adversarial examples than state-of-the-art targeted attack methods in deep hashing based retrieval.
\end{itemize}

\section{Related Work}
\subsection{Deep Hashing based Similarity Retrieval}

Existing deep hashing methods can be roughly grouped into unsupervised deep hashing and supervised deep hashing. Unsupervised deep hashing methods learn deep features of samples by preserving the structure or metric consistence embedded in samples without using any semantic labels, which are usually achieved by unsupervised representation learning \cite{salakhutdinov2009semantic, shen2018unsupervised, ghasedi2018unsupervised}.
Although the unsupervised schemes are more general, their performance for retrieval is not satisfactory because of the semantic gap dilemma \cite{smeulders2000content}.
Supervised deep hashing methods use the class labels or pairwise similarities as the semantic supervision in the learning process, yielding promising results \cite{xia2014supervised, lai2015simultaneous, zhu2016deep, li2016feature, liu2016deep, cao2017hashnet, jiang2017asymmetric, cao2018deep, zhang2020inductive}. For example, the first deep hashing method \cite{xia2014supervised} separates the whole hash learning into two steps: hash codes learning and data encoding. Recent works \cite{lai2015simultaneous, zhu2016deep, li2016feature} validated the importance of jointly learning similarity-preserving hash codes and minimizing the quantization error in continuous-binary space transformation, and also showed the nonlinear deep hashing function learning could greatly improve the retrieval performance in an end-to-end training architecture.
Thanks to the power of deep learning, researchers have extended the above methods to other complex tasks, \eg, \cite{jiang2017deep, li2018self, zhang2018binary}.

\subsection{Adversarial Attacks}
In image classification, an adversarial example is usually a carefully modified image, which is intentionally perturbed by adding a visually imperceptible perturbation to the original image but can confuse the deep model to misclassify it. Since Szegedy \textit{et al.} \cite{szegedy2013intriguing} discovered the properties of adversarial examples, various adversarial attack methods in image classification have been proposed to fool a trained DNN. According to the information of target model exposed to the adversary, adversarial attacks can be categorized as white-box attacks (\eg, FGSM \cite{goodfellow2014explaining}, I-FGSM \cite{kurakin2016adversarial}, PGD \cite{madry2017towards} and C\&W \cite{carlini2017towards}) and black-box attacks (\eg SBA \cite{papernot2017practical} and ZOO \cite{chen2017zoo}).
For white-box attack, the adversary knows the whole network architecture and parameters so that it can design the adversarial perturbations by calculating the gradient of the loss \textit{w.r.t.} inputs.
As for black-box attack, only the input and the output are available to the adversary, thus it is more challenging and practical.
However, these attack methods are optimization-based and they are quite slow for accessing the target model many times.
Recently, generation-based attack methods (\eg, \cite{baluja2017adversarial, xiao2018generating, goodfellow2014generative, mopuri2018nag, poursaeed2018generative, han2019once}) received much more attention due to their high-efficiency during test phase.
Generation-based attack methods learn a generative model which transforms the input images into the adversarial samples.
Once the generative model trained, it do not need to access the target model again and can generate adversarial examples with one-forward pass.


In addition to image classification, recent works on similarity retrieval \cite{li2019universal, tolias2019targeted, feng2020adversarial, yang2018adversarial, bai2020targeted, li2019cross, li2020vulnerability} have also confirmed the vulnerability of DNNs to adversarial examples.
Currently, there are only two works on attacking deep hashing models, \textit{i.e.}, \cite{yang2018adversarial} for non-targeted attack and \cite{bai2020targeted} for targeted attack. Specifically, HAG \cite{yang2018adversarial} is to make the hash code of the adversarial example as dissimilar as possible from that of the original example.
Bai \etal \cite{bai2020targeted} proposed the two targeted attack schemes in hashing based retrieval, dubbed \textit{point-to-point} (P2P) and \textit{deep hashing targeted attack} (DHTA).
P2P randomly chooses a hash code of a sample with the target label to direct the adversarial example by maximizing their similarity of hash codes.
DHTA transforms the targeted attack into a \textit{point-to-set} optimization problem, which maximizes the similarity between the hash code of the adversarial example and the set of hash codes of samples with the target label. In detail, DHTA selects the \textit{anchor code} \cite{bai2020targeted} which has the smallest distance to the set as target code to guide the optimization of the adversarial example by minimizing the Hamming distance between the hash code of the adversarial example and the anchor code.
Although making some progress in adversarial attacks of deep hashing, the existing works are generalized built on heuristic rules and optimization-based attack, which are inefficient and less effective.
In this work, we, for the first time, design a neural network (PrototypeNet) to learn the prototype code of the target label for targeted attack and use a generative model to achieve the entire attack framework.

\section{Generation-based Hashing Targeted Attack}
We propose a novel prototype-supervised adversarial network (ProS-GAN) to efficiently generate adversarial examples for targeted attack of deep hashing-based image retrieval.
Given a query image and a target label, targeted attack aims to learn an adversarial example for the query, whose nearest neighbors retrieved by the target hashing model from the database are semantically relevant to the target label.
As shown in Figure \ref{fig:framework}, the overall framework is a generative adversarial network \cite{goodfellow2014generative} and includes three sub-networks: a well-designed PrototypeNet $P$ for learning representative embedding of target labels, an adversarial generator $G$ and a discriminator $D$ for generating adversarial examples.
Specifically, $P$ embeds the target label into the semantic representation and outputs the predicted label and the corresponding prototype code which can be used to supervise the generation of adversarial examples.
$G$ learns to transform the given query image into the targeted adversarial example.
$D$ aims at distinguishing the generated image from the real one and categorizing them into the target and the original category, respectively.
$G$ and $D$ are trained in an adversarial manner, which encourages $G$ to generate more realistic images and ensures the semantic representation informative.

\subsection{Problem Formulation}
Let $O=\{(x_i, y_i)\}_{i=1}^N$ denote a dataset containing $N$ instances labeled with $C$ classes, where $x_i$ indicates the original image for the $i$-th instance, and $y_i=[y_{i1},...,y_{iC}] \in \{0, 1\}^C$ corresponds to a multi-label vector. $y_{ij}=1$ indicates that $x_i$ belongs to class $j$. Let $L=\{y_i\}_{i=1}^M$ denote all unique label dataset from $O$, where $M$ is the number of labels and $M \leq N$. We use similarity matrix $S$ to describe semantic similarities between each pair of data points. For any two instances $x_i$ and $x_j$, $S_{ij}=1$ indicates they share as least one label, otherwise $S_{ij}=0$. Similarly, we can use $S_{tj}=1$ to indicate that a label $y_t$ and a instance $(x_j,y_j)$ have similar semantics.

Hashing aims to transform semantically similar data items into similar binary codes for efficient nearest neighbor search \cite{wang2017survey}.
For a given deep hashing model $F(\cdot)$, the hash code of the sample $x_i$ is generated by
\begin{equation}
	\begin{aligned}
		\label{eq1-plus}
		& b_i = F(x_i) = \operatorname{sign}(f_\theta(x_i)), \quad \text{s.t. }  b_i\in\{-1,1\}^K,
	\end{aligned}
\end{equation}
where $f(\cdot)$ is a DNN with parameters $\theta$ to approximate $F(\cdot)$, $\operatorname{sign}(\cdot)$ is the $\operatorname{sign}$ function which binarizes the output of $f_\theta(\cdot)$ to $-1$ or $1$, and $K$ is the hash code length.
We use $B=(b_1 \ b_2 \ ... \ b_N)_{K \times N}$ to represent the hash code matrix for $O$.
In general, $f(\cdot)$ is a convolutional neural network (CNN) \cite{krizhevsky2012imagenet, simonyan2014very, he2016deep}, which consists of a convolutional feature extractor followed by fully-connected layers.
In particular, deep hashing methods adopt $\operatorname{tanh}(\cdot)$ function to approximate the $\operatorname{sign(\cdot)}$ function during training process.

In image retrieval, given a benign query image $x$ and a target label $y_t$, the goal of targeted attack is to generate corresponding adversarial example $x^\prime$, which could cause the target model to retrieve the images semantically related to the target label.
In addition, the adversarial perturbations (\textit{i.e.}, $x^{\prime}-x$) should be as small enough to be imperceptible to human eyes.
In this paper, we aim to design a function $\Phi$ to achieve such objective, \textit{i.e.},
\begin{equation}
	\begin{array}{c}
		\Phi:(x,y_t) \rightarrow x^\prime, \\
		\text{ s.t. }
		\min \sum_{i} d(F(x^\prime), F(x_i^{(t)})) - \sum_{j} d(F(x^\prime), F(x_j^{(n)})), \\
		\|x-x^\prime\|_p\leq\epsilon,
	\end{array}
	\label{eq:task}
\end{equation}
where $d(\cdot, \cdot)$ is a distance measure, $\|\cdot\|_p$ ($p={1,2,\infty}$) denotes $L_p$ norm, and $\epsilon$ is the maximum magnitude of adversarial perturbations.
$x_i^{(t)}$ is a sample semantically relevant to the target label, and $x_j^{(n)}$ is a irrelevant sample.
The minimized objective in Eqn. (\ref{eq:task}) ensures the hash code of the adversarial example $x^\prime$ as close as possible to those of the semantically relevant samples, and simultaneously stays away from those of semantically irrelevant ones.

\subsection{Prototype Generation}
Unlike targeted attack in image classification, deep hashing models aim to generate semantic-preserving hash codes instead of categories, and thus labels can not directly used for guiding the generation of targeted adversarial samples. In hashing based retrieval, the most intuitive idea for targeted attack is that we can construct the most representative hash code of samples with the target label, and then use it to supervise the learning process of the adversarial example generation. As such, we construct a semantic encoding strategy, \textit{i.e.}, PrototypeNet, to produce the prototype codes, which are used for representing the target labels. In PrototypeNet, the semantic representations are transformed into the corresponding prototype codes, and meanwhile could preserve the category knowledge of each target label.

Let $\theta_{p}$ denote the network parameters of the PrototypeNet $P$, and the objective function is defined as follows:
\begin{equation}
\begin{aligned}
\min_{\theta_{p}} \mathcal{L}_{pro} &= \alpha_1 \mathcal{J}_{1}+ \alpha_2 \mathcal{J}_{2}+\alpha_3 \mathcal{J}_{3} \\
&=-\alpha_1 \sum_{i=1}^{M}\sum_{j=1}^{N}\left(S_{i j} \Omega_{i j}-\log \left(1+e^{\Omega_{i j}}\right)\right) \\
&+\alpha_2\left\|H-B^{(p)}\right\|_{F}^{2}+\alpha_3\left\|\hat{Y}-Y\right\|_{F}^{2}, \\
& \text { s.t.} \quad B^{(p)} \in\{-1,1\}^{K\times M},
\end{aligned}
\label{eq:obj_p}
\end{equation}
where $S$ is the semantic similarity matrix between the target labels and image instances from $O$, $\Omega_{ij}=\frac{1}{2}(H_{*i})^{T}(B_{*j})$, and $B$ is the hash code matrix for $O$. $H$ is the predicted hash codes for the targeted labels $Y$, and $\hat{Y}$ are the predicted labels. $B^{(p)}$ is the expected binary codes of $H$, \ie, $B^{(p)}=\operatorname{sign}(H)$. $\alpha_1$, $\alpha_2$, $\alpha_3$ are hyper-parameters. $\|\cdot\|_F$ denotes Frobenius norm.

The first term $\mathcal{J}_{1}$ in (\ref{eq:obj_p}) is the negative log-likelihood of the pair-wise similarity in $S$. Given $S$, the probability of $S$ under the condition $B$ can be
defined as follows:
\begin{equation}
\begin{aligned}
p\left(S_{i j} \mid B\right)=\left\{\begin{array}{ll}
\sigma\left(\Omega_{ij}\right), & S_{ij}=1 \\
1-\sigma\left(\Omega_{ij}\right), & S_{ij}=0
\end{array}\right.
\end{aligned}
\end{equation}
where  $\sigma\left(\Omega_{i j}\right)=\frac{1}{1+e^{-\Omega_{i j}}}$.
Notably, this pair-wise class encoding process can maximally capture the category information of the target label.
Moreover, by using the above pair-wise similarity preservation loss in $\mathcal{J}_{1}$, the prototype codes can jointly maximize the compactness with the hash codes from semantically-relevant samples and separability with those from semantically-irrelevant samples.
Hence, by optimizing $\mathcal{J}_{1}$, the generated prototype codes can maintain the representative semantics of the target labels and the discriminative characteristics.

$\mathcal{J}_{2}$ is the quantization loss to minimize the approximation error between the prototype embedding $H$ and the expected binary codes $B^{(p)}$. $\mathcal{J}_{3}$ is the classification loss of the semantic representation $r_t$ to keep its category information.

\subsection{Adversarial Generator $G$}
Given a semantic representation $r_t$ from PrototypeNet and an original image $x$, we design an adversarial generator $G$ to learn the targeted adversarial example of $x$. Particularly, we integrate the decoded semantic representation and the original image into a well-designed encoder-decoder network with skip connection strategy. 
Generally, it is mainly composed of two parts: a semantic representation decoder $D_t$ and a image encoder-decoder $G_{xt}$. $D_t$ is used to upsample $r_t$ to $x_t$ with the same size of $x$. Then, $x$ and $x_t$ are concatenated into $G_{xt}$ to generate the adversarial example. Inspired by the skip connection in \cite{ronneberger2015u, zhu2018hidden}, we concatenate the original image $x$ and the output of the last deconvolutional layer in $G_{xt}$, which can facilitate the reconstruction of the adversarial example $x^{\prime}$ during training.

To guarantee the high-quality of the adversarial examples, we define the objective of the generator $G$ as follows:
\begin{equation}
\begin{aligned}
\min_{\theta_g} \mathcal{L}_{gen} = \sum_{y_t \in L, (x, y) \in O} ( \mathcal{J}_{ham} + \alpha\mathcal{J}_{re} + \beta \mathcal{J}_{adv} ),
\end{aligned}
\label{eq:g}
\end{equation}
where $\alpha$, $\beta$ are the weighting factors, and $\theta_g$ is the parameters of $G$.
$\mathcal{J}_{ham}$ is the Hamming distance loss, $\mathcal{J}_{re}$ is the reconstruction loss and $\mathcal{J}_{adv}$ is the adversarial loss.

$\mathcal{J}_{ham}$ is the Hamming distance loss that enforces the hash code of the adversarial example similar to the hash codes of samples relevant to the target label.
Since we take the prototype code as representative of the target label, we can choose the prototype code as a target code to guide the generation of the adversarial example. As such, we can directly minimize the Hamming distance between the hash code of the adversarial example and the prototype code:
\begin{equation}
\begin{aligned}
\mathcal{J}_{ham} = d_H(h_{x^\prime}, h_t),
\end{aligned}
\end{equation}
where $d_H(\cdot, \cdot)$ is the Hamming distance operator, $h_{x^{\prime}}$ is the hash code of $x^{\prime}$, and $h_t$ is the prototype code of the target label $y_t$.
Due to $d_H(h_i, h_j)=\dfrac{1}{2}(K-h_i^Th_j)$, we can replace Hamming distance with inner product.
Besides, we normalize the Hamming distance to range $[0,2]$.
Therefore, the final $\mathcal{J}_{ham}$ for instance $x$ is calculated as follows:
\begin{equation}
\begin{aligned}
\mathcal{J}_{ham} =& -\frac{1}{K}h_t^T f(x^\prime)+1 \\
=& -\frac{1}{K}h_t^T f(G(x,r_t)) + 1,
\end{aligned}
\end{equation}
where $h_{x^\prime}$ is approximated by $f(x^\prime)$, and $r_t$ is the semantic representation of $y_t$ produced by PrototypeNet $P$.

$\mathcal{J}_{re}$ is the reconstruction loss that ensures the pixel difference between the adversarial example and the original image as small as possible, \textit{i.e.}, the adversarial perturbations are small enough to be imperceptible.
We simply adopt $L_2$-norm loss to measure the reconstruction error as follows:
\begin{equation}
\begin{aligned}
\mathcal{J}_{re}
= \|x-x^{\prime}\|_2^2 = \|x-G(x, r_t)\|_2^2.
\end{aligned}
\end{equation}

$\mathcal{J}_{adv}$ is the adversarial loss that ensures the semantic representation informative to enhance attack performance and encourages the generated adversarial example looks real. We re-formulate the objective label of $\mathcal{J}_{adv}$ as follows:
\begin{equation}
\begin{aligned}
\widetilde{y_t} = \underbrace{[y_{t1},y_{t2},...,y_{tC}}_{y_t},0],
\end{aligned}
\end{equation}
where $y_t$ is in one-hot encoding, and the last node in $\widetilde{y_t}$ is for the fake sample.
Hence, $\mathcal{J}_{adv}$ is defined as follows:
\begin{equation}
\begin{aligned}
\mathcal{J}_{adv} = \|D(x^\prime)-\widetilde{y_t}\|_2^2.
\end{aligned}
\end{equation}

\begin{table*}[!t]
	\caption{t-MAP (\%) of targeted attack methods and MAP (\%) of benign samples for different code lengths on three datasets.}
	\begin{center}
		\setlength{\tabcolsep}{1.0mm}
		\resizebox{0.99\textwidth}{!}{
			\begin{tabular}{lc|cccc|cccc|cccc}
				\hline
				\multirow{2}{*}{Method} & \multirow{2}{*}{Metric} & \multicolumn{4}{c|}{FLICKR-25K}       & \multicolumn{4}{c|}{NUS-WIDE}         & \multicolumn{4}{c}{MS-COCO}              \\ \cline{3-14}
				&                 & 12 bits & 24 bits & 32 bits & 48 bits & 12 bits & 24 bits & 32 bits & 48 bits & 12 bits & 24 bits & 32 bits & 48 bits \\ \hline
				Original         & t-MAP    & 63.58 & 63.49 & 63.49 & 63.59 & 55.51 & 55.57 & 55.67 & 55.86 & 42.33 & 42.60 & 42.67 & 42.90  \\
				Noise            & t-MAP    & 63.37 & 63.40 & 63.45 & 63.55 & 54.23 & 54.59 & 55.61 & 55.83 & 42.33 & 42.60 & 42.67 & 42.90  \\
				P2P  & t-MAP & 82.55 & 83.79 & 84.65 & 84.44 & 70.33 & 71.44 & 71.98 & 73.26 & 56.77 & 59.11 & 59.87 & 59.72 \\
				DHTA & t-MAP & 86.27 & 87.74 & 88.35 & 88.48 & 74.04 & 75.52 & 75.65 & 75.93 & 59.85 & 61.90 & 63.22 & 63.20 \\
				ProS-GAN         & t-MAP    & \textbf{89.05} & \textbf{89.85} & \textbf{91.10} & \textbf{91.09} & \textbf{77.73} & \textbf{78.21} & \textbf{78.25} & \textbf{78.75} & \textbf{66.22} & \textbf{71.28} & \textbf{71.65} & \textbf{68.92} \\ \hline
				Anchor code \cite{bai2020targeted}     & t-MAP    & 86.40 & 87.98 & 88.68 & 88.97 & 75.15 & 77.41 & 78.40 & 78.12 & 60.35 & 63.14 & 64.41 & 64.65  \\
				Prototype code  & t-MAP    & 91.33 & 92.20 & 92.91 & 93.12 & 79.04 & 80.71 & 81.34 & 81.50 & 67.38 & 71.30 & 71.93 & 69.98 \\ \hline
				Original         & MAP      & 78.88 & 80.12 & 80.75 & 80.87 & 69.54 & 70.80 & 71.33 & 71.21 & 57.64 & 61.02 & 62.63 & 62.98 \\ \hline
			\end{tabular}
		}
	\end{center}
	\label{tab:results}
	\vspace{-0.3cm}
\end{table*}

\begin{figure*}[!t]
	\centering
	{\includegraphics[width=1.1in]{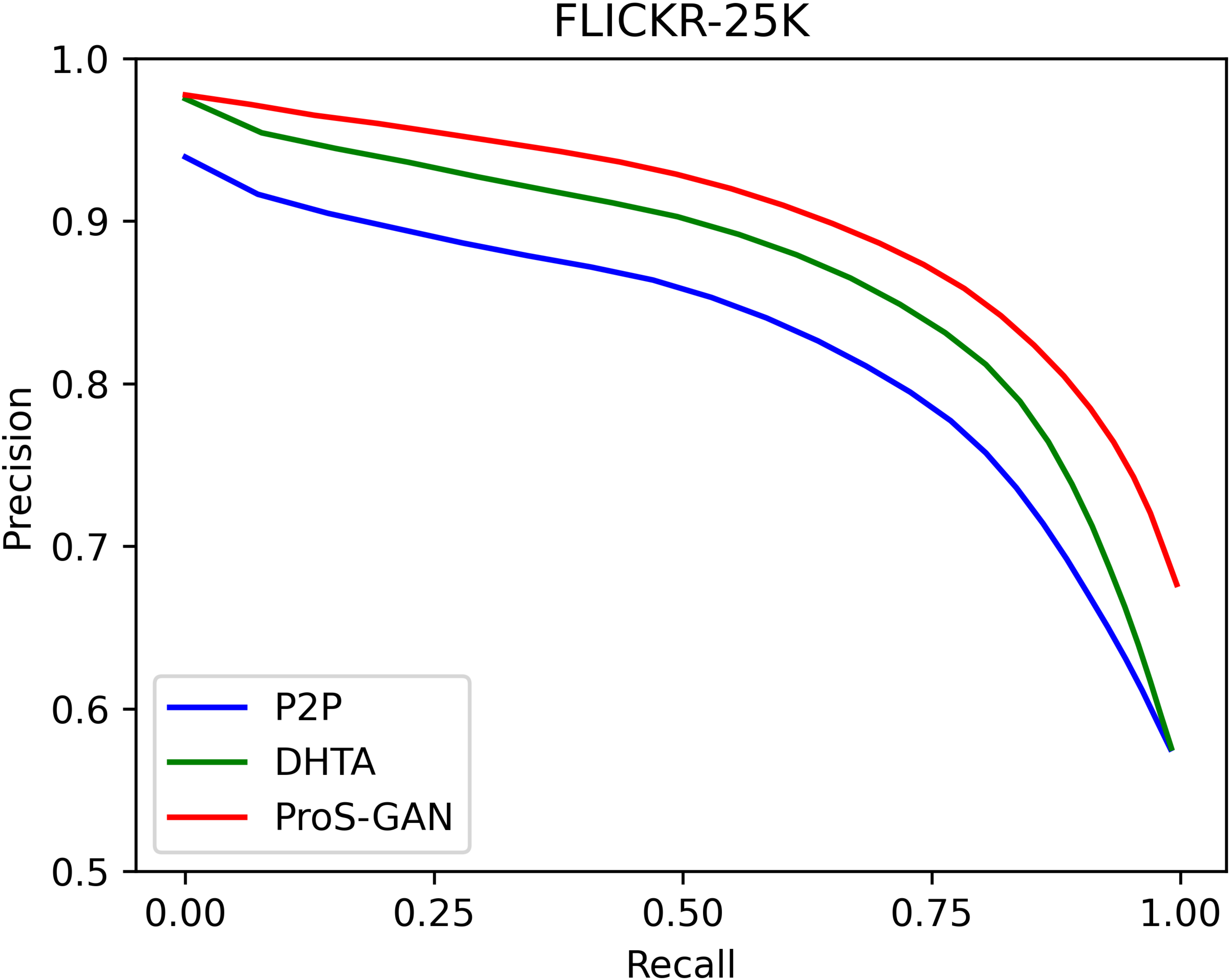}}
	{\includegraphics[width=1.1in]{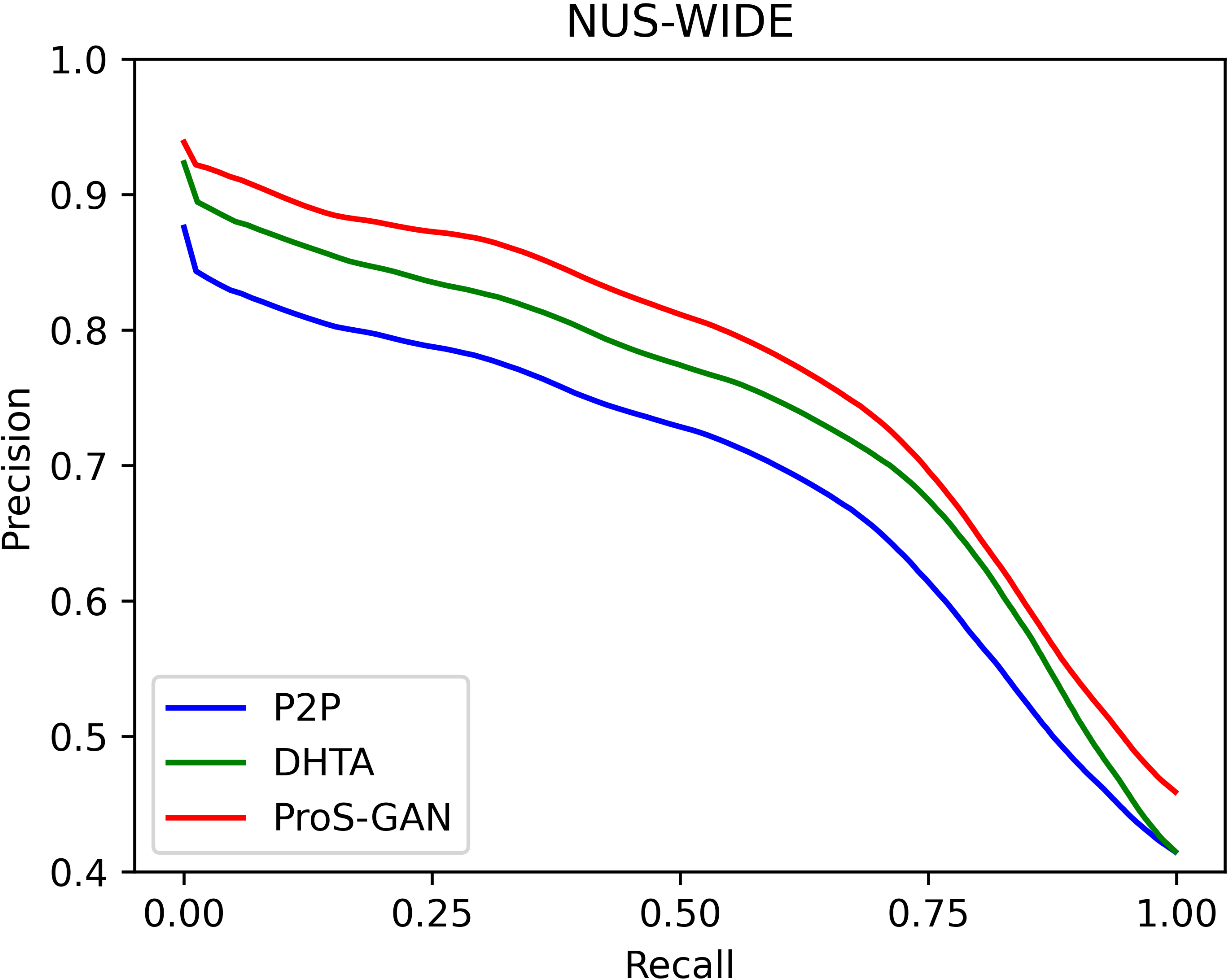}}
	{\includegraphics[width=1.1in]{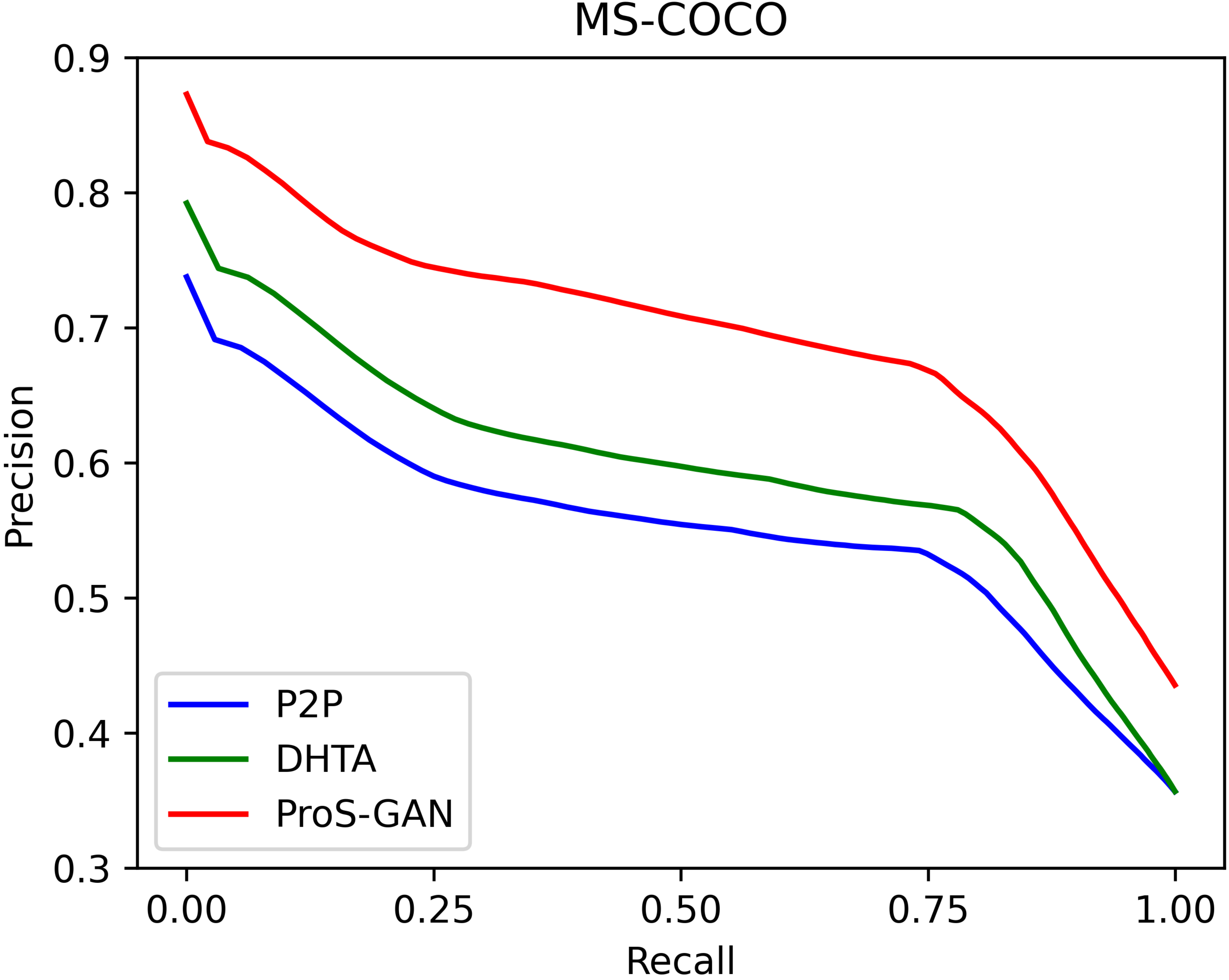}}
	{\includegraphics[width=1.1in]{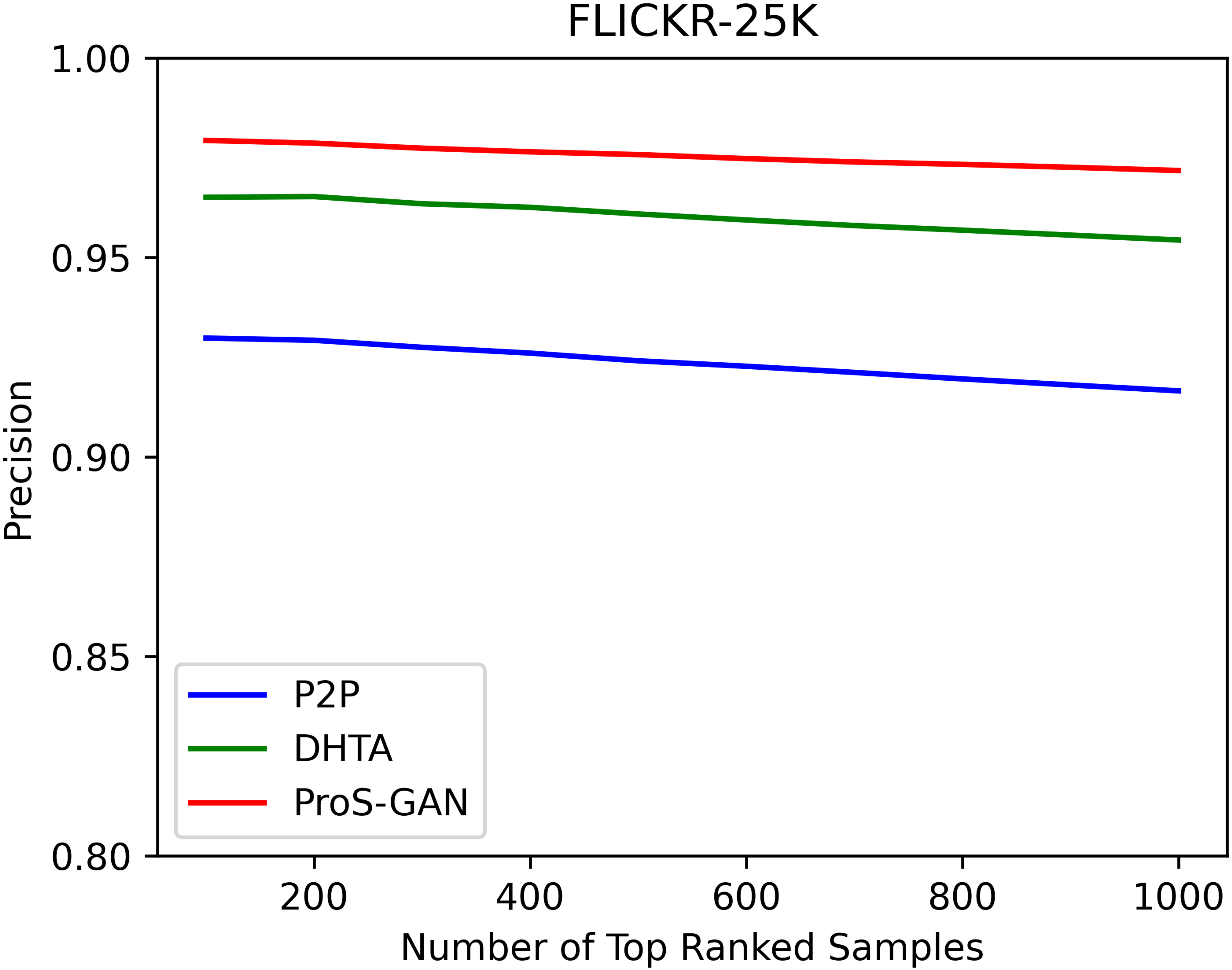}}
	{\includegraphics[width=1.1in]{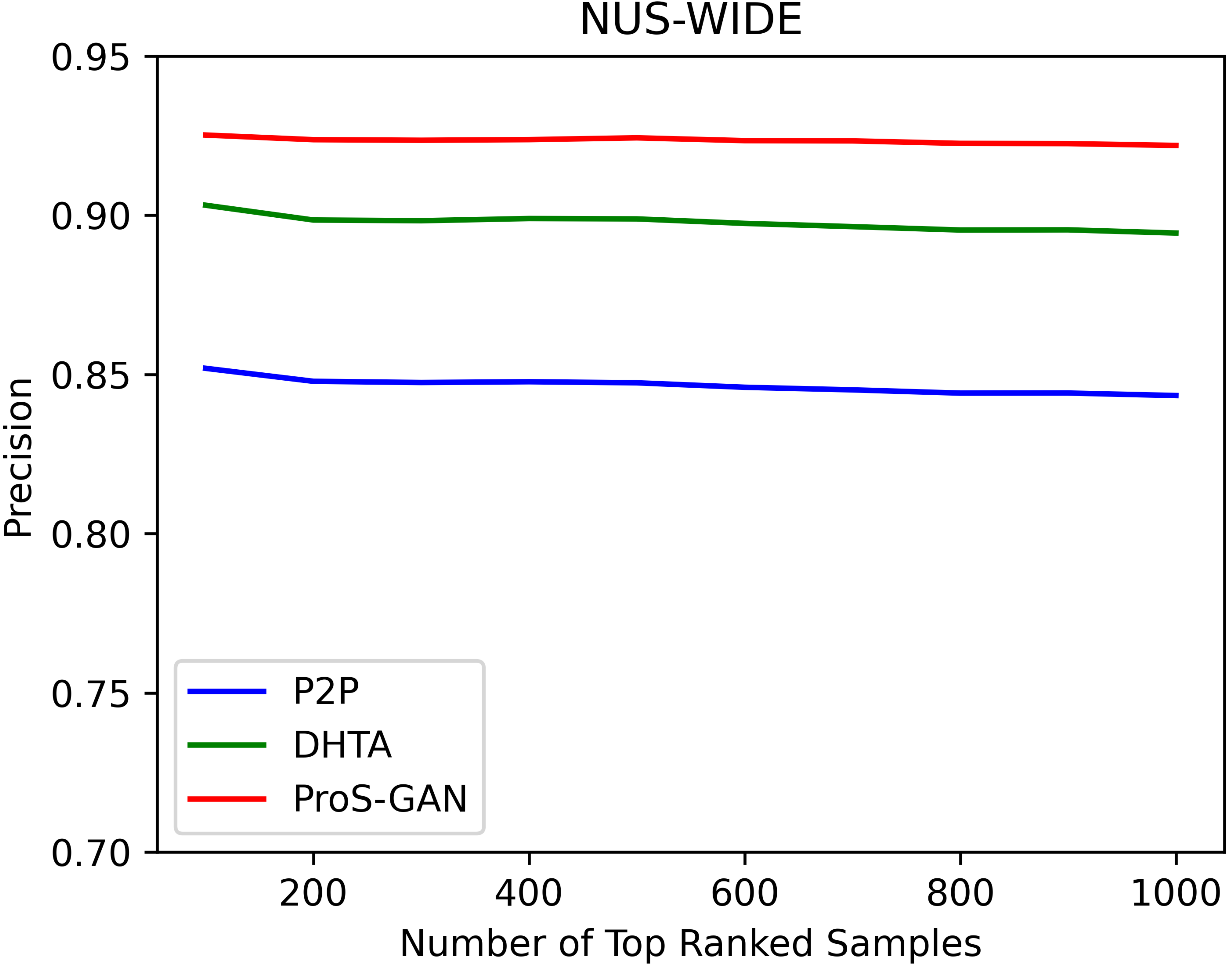}}
	{\includegraphics[width=1.1in]{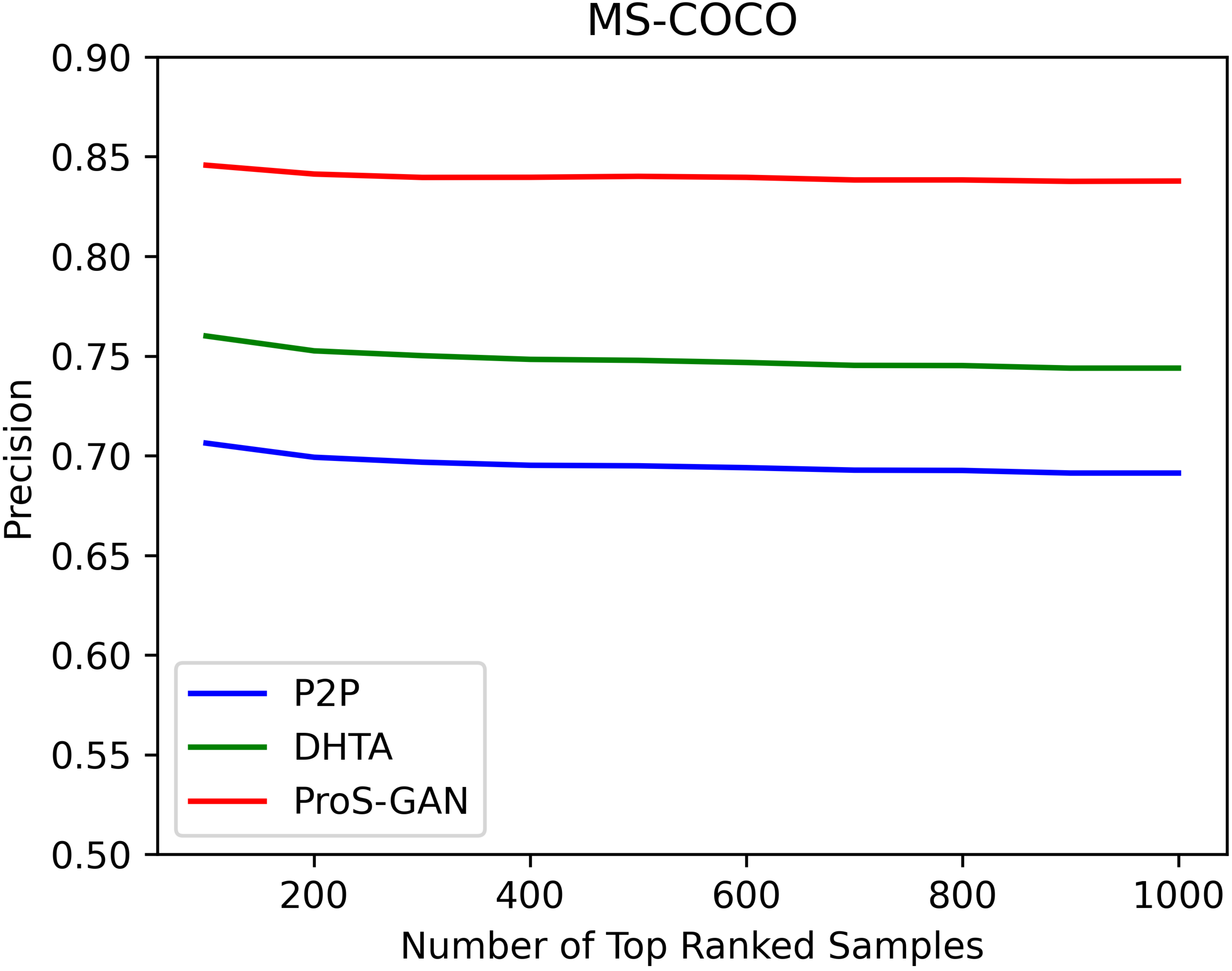}}
	\caption{Precision-Recall and precision@topN curves on three datasets under 32 bits code length.}
	\label{fig:pr}
	\vspace{-0.1cm}
\end{figure*}

\subsection{Discriminator $D$}
The designed discriminator is to distinguish the fake images (\textit{i.e.}, the adversarial images) from the real images (\textit{i.e.}, the benign images) and to categorize them into the target categories and the original categories, respectively.
This strategy has two advantages: on one hand, the adversarial learning between the $G$ and $D$ encourages the generated adversarial examples look more realistic; on the other hand, the discriminator for category classification can ensure the representation $r_t$ semantically informative, which can enforce the category information of the representation to be embedded into the generated images and further improve the targeted attack performance.
Specifically, the output of $D$ is a Sigmoid layer with $C+1$ nodes in order to predict the category and to distinguish real/fake together, where the first $C$ nodes and the last one indicate the category and the falsity for the input image, respectively.

By inputting the real image $x$, the objective label for the discriminator is re-formulated as follows:
\begin{equation}
\begin{aligned}
\widetilde{y}= \overbrace{[y_{1},y_{2},...,y_{C}}^{y},0],
\end{aligned}
\end{equation}
where $y$ is the label of the original image $x$. We set $(C+1)$-th node as $0$ for real samples. For the fake image, the objective label is re-formulated as follows:
\begin{equation}
\begin{aligned}
\widetilde{y_t} = \overbrace{[y_{t1},y_{t2},...,y_{tC}}^{y_t},1],
\end{aligned}
\end{equation}
where $y_t$ is the target label, and we set $(C+1)$-th node as $1$ for fake samples.
In summary, the objective loss function of $D$ is formulated as below:
\begin{equation}
	\begin{aligned}
		\min_{\theta_d} \mathcal{L}_{dis} = \sum_{y_t \in L, (x, y) \in O} \frac{1}{2} \left( \|{D(x)-\widetilde{y}}\|_2^2 + \|{D(x^\prime) - \widetilde{y_t}}\|_2^2 \right),
	\end{aligned}
\end{equation}
where $\theta_d$ is the parameters of $D$.

\section{Experiments}
\subsection{Datasets}
We evaluate our attack method on three popular multi-label datasets, \textit{i.e.}, \textbf{FLICKR-25K} \cite{huiskes2008mir}, \textbf{NUS-WIDE} \cite{chua2009nus} and \textbf{MS-COCO} \cite{lin2014microsoft}. \textit{\textbf{FLICKR-25K}} contains $25,000$ images with $38$ classes. Follwing \cite{wang2020deep}, we randomly sample $1,700$ images as queries, and the remaining as a database. Besides, we randomly select $5,000$ images from the database to train hashing models and our framework. \textit{\textbf{NUS-WIDE}} consists of $269,648$ images in $81$ categories. We only select $195,834$ images comprising the $21$ most frequent concepts. Following \cite{jiang2017deep}, we take $2,100$ images as a query set, and the rest samples as a database. Moreover, we sample $10,500$ images from the database as a training set. \textbf{\textit{MS-COCO}} contains $82,783$ training images and $40,504$ validation images, where each image is labeled with $80$ categories. We combine the training and validation sets, obtaining $122,218$ images. Following \cite{cao2017hashnet}, we randomly sample $5,000$ images as queries, and the rest regarded as a database. $10,000$ images are randomly sampled from the database as training points.

\subsection{Evaluation setup}
For image hashing, we select the objective function of DPSH \cite{li2016feature} as default method, which is one of the most representative deep hashing methods, to construct the target hashing model. Importantly, for any other popular deep hashing method, similar results could be achieved by our ProS-GAN. Specifically, VGG-11 \cite{simonyan2014very} is adopted as the default backbone network. We replace the last fully connected layer of VGG-11 with the hashing layer, including a new fully-connected layer and the $\operatorname{Tanh}$ activation.

For the network architecture, we built PrototypeNet with four-layer fully-connected networks ($y_t \rightarrow 4096 \rightarrow 512 \rightarrow r_t \rightarrow y_t, h_t$). We adopt a fully-connected layer and four deconvolutional layers for the Decoder $D_t$ to upsample the semantic representation $r_t$. We adapt the architecture for $G_{xt}$ from \cite{zhu2017unpaired}, and the discriminator contains five stride-2 convolutions and last layer with a $7 \times 7$ convolution.

After training ProS-GAN, we will use the PrototypeNet and the generator to attack the target hashing network. We set $\alpha_1$, $\alpha_2$ and $\alpha_3$ as $1$, $10^{-4}$ and $1$, respectively. The weighting factor $\alpha$ are set with $50$ for NUS-WIDE and MS-COCO, $100$ for FLICKR-25K, and $\beta$ is set as $1$. We train ProS-GAN using Adam \cite{kingma2014adam} optimizer with initial learning rate $10^{-4}$. The training epochs are $100$ in batch size $24$. The ProS-GAN is implemented via PyTorch and is run on NVIDIA TITAN RTX GPUs. For the optimization procedure of ProS-GAN, please refer to the supplementary file.

Following \cite{bai2020targeted}, we adopt t-MAP (targeted mean average precision) to evaluate the targeted attack performance, instead of MAP (mean average precision). Since t-MAP uses the target labels as the test labels, the higher the t-MAP, the stronger the targeted attack. In image retrieval, we calculate t-MAP on all retrieved images from database. Besides, we also present the precision-recall curves (PR curves) and precision@topN curves. In detail, we randomly select a label as target label for each generation of adversarial examples. We compare the proposed framework with gradient-based methods, including P2P \cite{bai2020targeted} and DHTA \cite{bai2020targeted}. For fair comparison, the experimental settings of P2P and DHTA are same as \cite{bai2020targeted}, where the perturbation magnitude $\epsilon$ is set to $8/255$.

\begin{figure}
	\begin{center}
		\includegraphics[width=0.99\linewidth]{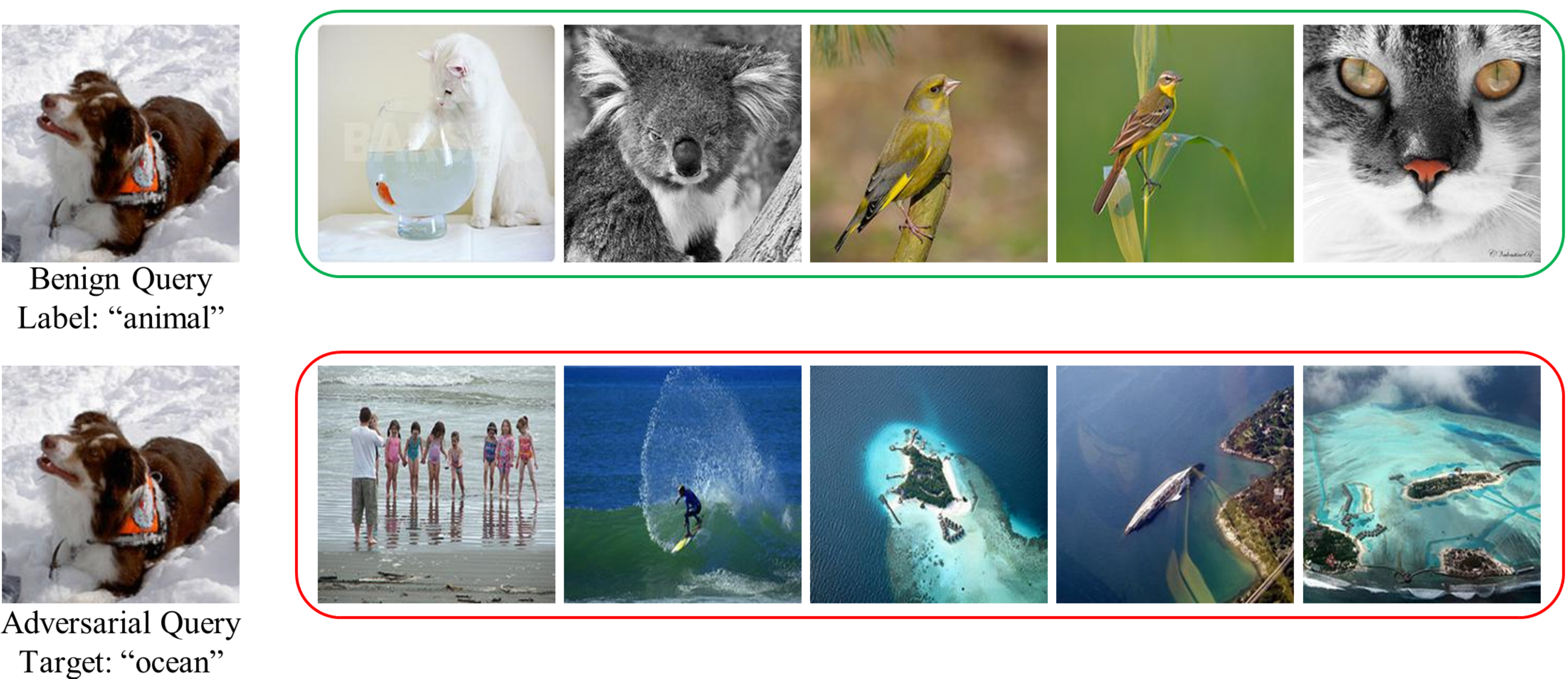}
	\end{center}
	\caption{An example to retrieve top-5 similarity samples on NUS-WIDE with the benign query and its adversarial query.}
	\label{fig:retrieval}
\end{figure}

\subsection{Results}
\textbf{Targeted Attack Performance:}
We provide the targeted attack performance of different methods, as shown in Table \ref{tab:results}, using t-MAP criteria for comparison.
The \textit{Noise} in Table \ref{tab:results} is to query with noisy samples which are benign images with additive noises sampled from the uniform distribution $U(-\epsilon, +\epsilon)$.
The t-MAP values of \textit{Noise} are almost the same as the values of querying with benign samples (called \textit{Original}) on FLICKR-25K, NUS-WIDE and MS-COCO datasets, which indicates the images with random noises can not bias the predictions of deep hashing models.
In contrast, all the t-MAP values of DHTA and ProS-GAN are higher than the t-MAP values of 'Original', which verifies the effectiveness of adversarial attacks.
Moreover, all the t-MAP values of our ProS-GAN are better than all the previous methods including P2P and DHTA.
For example, compared with the state-of-the-art DHTA, we achieve absolute boosts of more than 2\% in t-MAP for various number of bits on both FLICKR-25K and NUS-WIDE.
On MS-COCO, our method outperforms DHTA over 5\% in all cases.
Especially, for 24 bits, the t-MAP of ProS-GAN is higher than DHTA by 9.38\% on MS-COCO.
As shown in Table \ref{tab:results}, our superior performance benefits from the superiority of the prototype code over the anchor code, which indicates the prototype code is a more representative code of the target label.
Furthermore, the targeted retrieval performance on three datasets in terms of the PR and precision@topN curves are shown in Figure \ref{fig:pr} for comprehensive comparison. The curves of ProS-GAN are always above all other curves of previous methods, which also shows our performance does surpass all other methods.
An example of the retrieval results with a benign image and its adversarial example generated by our method is displayed in Figure \ref{fig:retrieval}.

\textbf{Perceptibility \& Efficiency:}
In addition to attack performance, the \textit{perceptibility} is also an important criteria to evaluate the quality of adversarial examples.
Following \cite{szegedy2013intriguing}, the perceptibility is calculated by $\sqrt{\frac{1}{Z}\|x^\prime-x\|_2^2}$, where $Z$ is the pixel number, and pixel values are all normalized in the range $[0,1]$.
The higher the perceptibility, the worse visual quality of adversarial examples.

To make comprehensive comparison between efficiency and perceptibility of adversarial examples generated by various methods, we record t-MAP, perceptibility and generating time for $32$-bits length on three datasets, which are summarized in Table \ref{tab:per_eff}. It is observed that ProS-GAN has the highest targeted attack performance, the second-best visual quality and the fastest generation speed for all datasets.
Specifically, DHTA with FGSM has fast speed to generate adversarial examples, but it has lower attack performance yet higher perceptibility.
On FLICKR-25K, ProS-GAN outperforms DHTA about $1660\times$ of generation speed.
Although ProS-GAN performs a little worse than DHTA \cite{bai2020targeted} on visual quality, DHTA needs multiple gradient descent to optimize adversarial perturbations and can not attack hashing models in real time. In summary, ProS-GAN not only outperform all the previous methods in attack performance and speed, but also can produce adversarial examples with high visual quality.

\begin{table}[!t]
	\caption{t-MAP (\%), perceptibility ($\times10^{-2}$) between benign samples and adversarial samples (per image) and generating time (second per image) on attacking hashing models with 32 bits length. The hyper-parameter settings of gradient-based attacks: for FGSM \cite{goodfellow2014explaining}, $\epsilon=8/255$; for I-FGSM \cite{kurakin2016adversarial}, $\epsilon=8/255$ and step size $\alpha=1/255$; for DHTA, the settings follows \cite{bai2020targeted}.}
	\setlength{\tabcolsep}{0.69mm}
	\begin{center}
		\resizebox{0.48\textwidth}{!}{
			\begin{tabular}{ll|ccc|ccc|ccc}
				\hline
				\multirow{2}{*}{Method} & \multirow{2}{*}{Iteration} & \multicolumn{3}{c|}{FLICKR-25K} & \multicolumn{3}{c|}{NUS-WIDE} & \multicolumn{3}{c}{MS-COCO} \\ \cline{3-11}
				&                            & t-MAP   & Per.   & Time   & t-MAP   & Per.   & Time  & t-MAP  & Per. & Time \\ \hline
				DHTA + FGSM    & 1      & 80.18   & 3.06   & 0.013   & 66.78   & 3.09   & 0.020   & 51.44  & 3.10 & 0.017   \\
				DHTA + I-FGSM  & 100    & 88.65   & 2.46   & 0.270   & 77.26   & 2.57   & 0.280   & 64.50  & 2.56 & 0.277   \\
				DHTA          & 2000   & 88.35   & \textbf{0.84}   & 9.957   & 75.65   & \textbf{0.80}   & 5.601   & 63.22  & \textbf{0.63} & 5.685   \\ \hline
				ProS-GAN       & 1      & \textbf{91.10}   & 2.23   & \textbf{0.006}   & \textbf{78.25}   & 1.87   & \textbf{0.005}   & \textbf{71.65}  & 1.86 & \textbf{0.005}   \\ \hline
			\end{tabular}
		}
	\end{center}
	\label{tab:per_eff}
\end{table}

\subsection{Ablation studies}
\begin{figure}[htbp]
	\centering
	\subfigure[]{
		\begin{minipage}[t]{0.403\linewidth}
			\centering
			\includegraphics[width=1.0\linewidth]{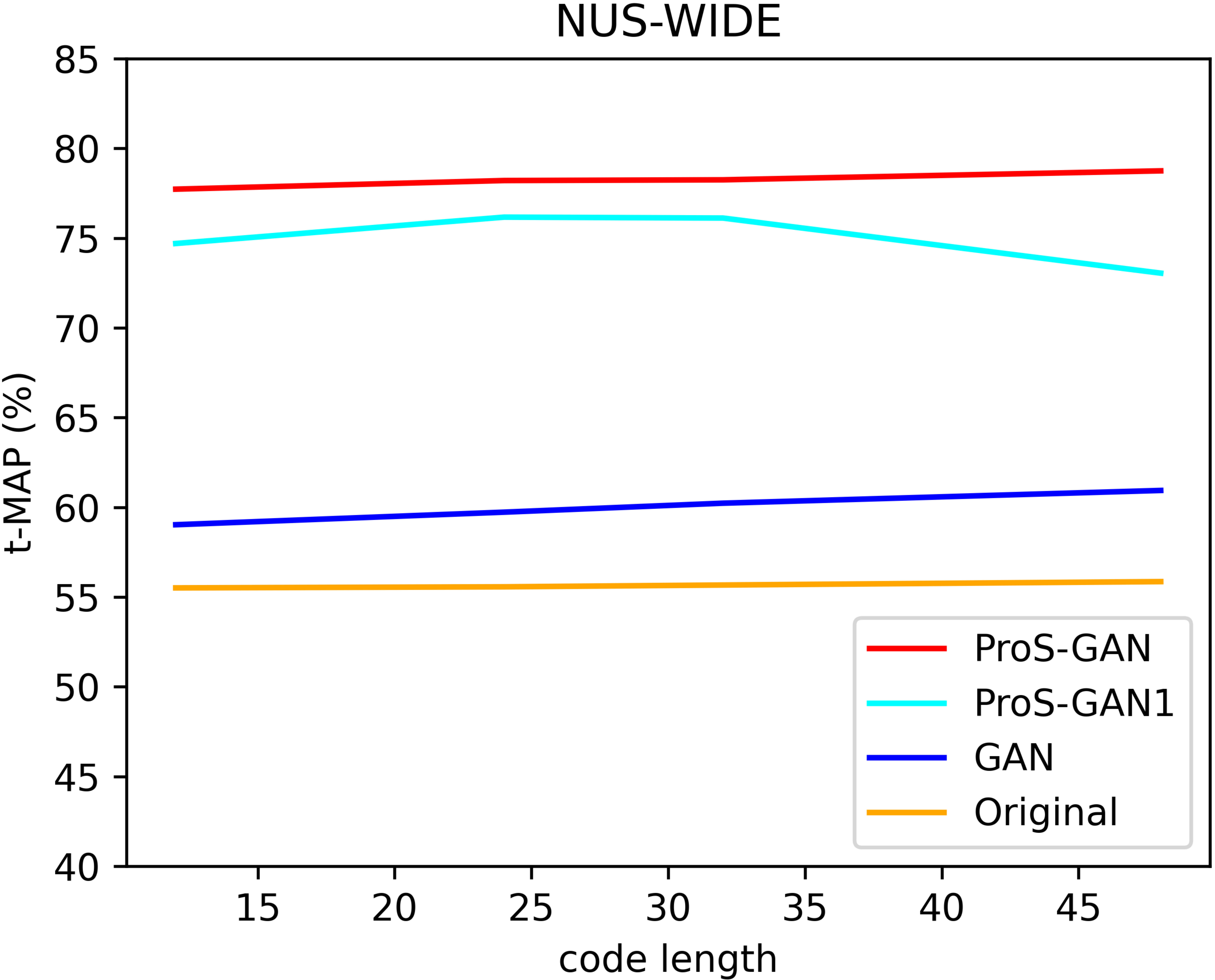}\\
		\end{minipage}%
		\label{fig:ablation}
	}%
	\subfigure[]{
		\begin{minipage}[t]{0.597\linewidth}
			\centering
			\includegraphics[width=0.8\linewidth]{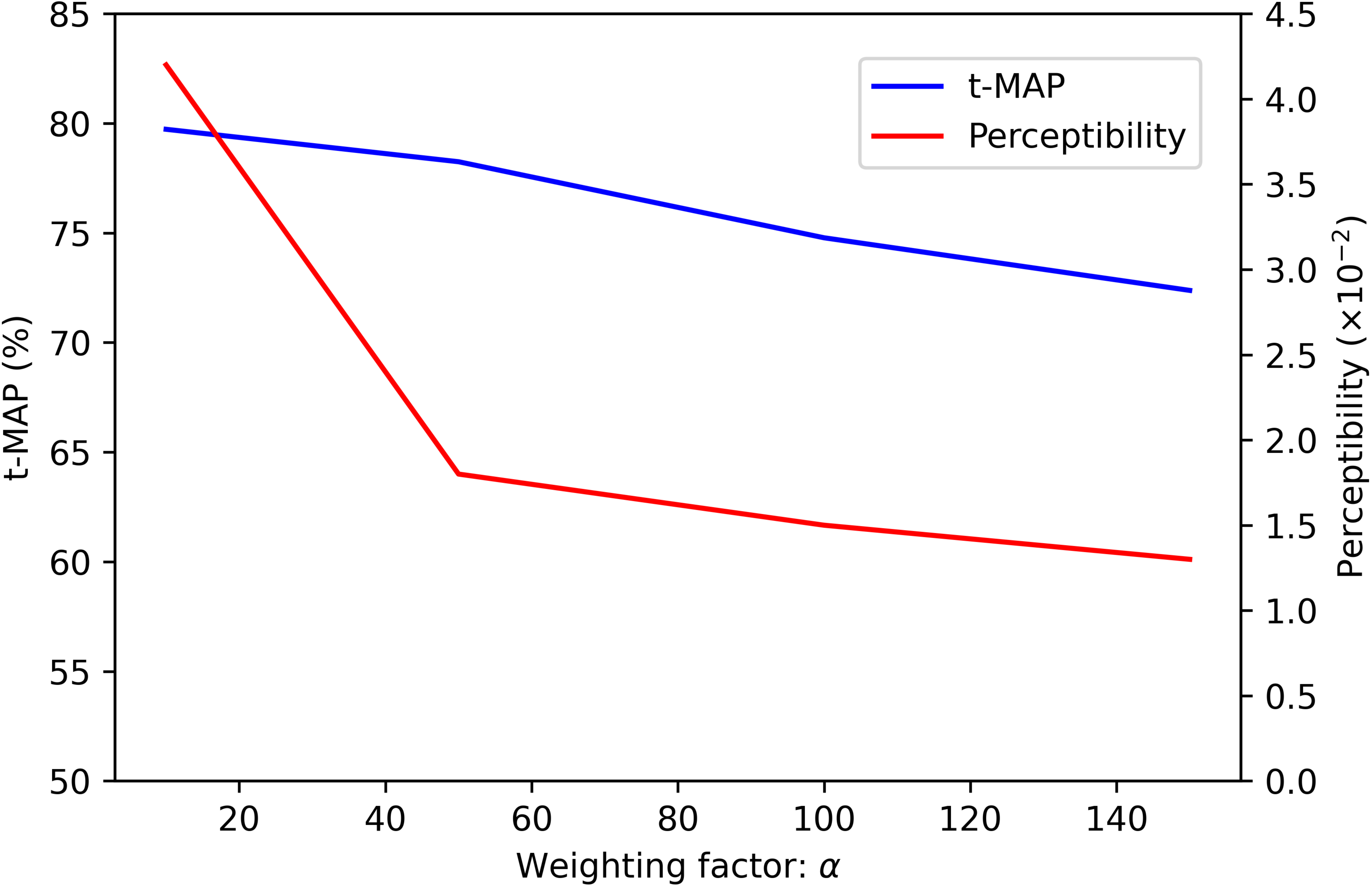}\\
		\end{minipage}%
		\label{fig:re_w}
	}%
	\centering
	\caption{(a) t-MAP (\%) for various code length on NUS-WIDE with different ablation architecture. (b) t-MAP (\%) and perceptibility ($\times10^{-2}$) for 32 bits code length on NUS-WIDE.}
	\label{fig:compare_fig}
\end{figure}

\textbf{Effect of the PrototypeNet:}
In order to explore the influence of the PrototypeNet for targeted attack performance, we remove the Hamming distance loss from the proposed structure denoted as \textit{GAN}.
As shown in Figure \ref{fig:ablation}, ProS-GAN outperforms GAN by a large margin.
Thus, the prototype code produced by PrototypeNet determines the performance of the targeted attack.

\textbf{Effect of the discriminator:}
In addition to making the generated adversarial examples look more realistic, we argue that the discriminator plays an important role in category classification and enforce the semantic representations informative, which can boost the attack performance.
In order to verify this point, we remove the classification module of the discriminator denoted as \textit{ProS-GAN1}, and the result is shown in Figure \ref{fig:ablation}.
The curve of ProS-GAN is above ProS-GAN1 in all different hash bits, which shows that the discriminator can indeed further improve the attack performance due to less interference to target information.

\textbf{Visual quality \textit{vs}. targeted retrieval precision:}
The weighting factor $\alpha$ controls the reconstruction quality of generated adversarial examples.
To explore the impact of different $\alpha$ on visual quality and attack performance of adversarial examples, we make comparison results with 32 bits on NUS-WIDE, as shown in Figure \ref{fig:re_w}.
When $\alpha$ increases, the visual quality gradually increases with the decreasing of perceptibility values,
but the attack performance gradually drops.
Thus, $\alpha$ can control the balance between the imperceptible quality and attack performance of adversarial perturbations.

\begin{table}[!t]
	\caption{Transfer t-MAP (\%) for the NUS-WIDE dataset. H-AlexNet, H-VGG11 and H-ResNet18 denote 12 bits DPSH models based on AlexNet \cite{krizhevsky2012imagenet}, VGG11 \cite{simonyan2014very} and ResNet18 \cite{he2016deep}, respectively, and "*" denotes their 32 bits variants.}
	\begin{center}
		\resizebox{0.48\textwidth}{!}{
			\setlength{\tabcolsep}{0.5mm}
			\begin{tabular}{ll|cccccc}
				\hline
				Method                    & Attacked model & H-AlexNet      & H-AlexNet*     & H-VGG11        & H-VGG11*       & H-ResNet18     & H-ResNet18*    \\ \hline
				\multirow{6}{*}{DHTA}     & H-AlexNet      & \textbf{71.11} & 70.39          & 56.25          & 57.09          & 53.64          & 52.89          \\
				& H-AlexNet*     & 68.27          & \textbf{71.86} & 55.98          & 56.98          & 53.36          & 52.54          \\
				& H-VGG11        & 54.94          & 55.11          & \textbf{74.04} & 74.94          & 54.78          & 54.48          \\
				& H-VGG11*       & 54.82          & 55.24          & 73.32          & \textbf{75.65} & 54.32          & 54.16          \\
				& H-ResNet18     & 54.11          & 54.56          & 54.69          & 55.51          & \textbf{67.55} & 66.38          \\
				& H-ResNet18*    & 54.03          & 54.46          & 54.31          & 55.41          & 65.34          & \textbf{70.08} \\ \hline
				\multirow{6}{*}{Ours} & H-AlexNet      & \textbf{75.13} & 74.96          & 63.70          & 64.14          & 58.51          & 58.43          \\
				& H-AlexNet*     & 73.81          & \textbf{78.03} & 64.35          & 65.94          & 62.13          & 63.12          \\
				& H-VGG11        & 60.69          & 60.75          & \textbf{77.73} & 76.05          & 62.20          & 61.99          \\
				& H-VGG11*       & 60.61          & 61.84          & 76.42          & \textbf{78.25} & 63.34          & 63.52          \\
				& H-ResNet18     & 59.06          & 59.35          & 60.59          & 60.50          & \textbf{70.79} & 69.69          \\
				& H-ResNet18*    & 59.25          & 58.89          & 59.36          & 59.76          & 66.92          & \textbf{75.21} \\ \hline
				\multicolumn{2}{c|}{Original}              & 54.09          & 54.45          & 55.51          & 55.67          & 53.54          & 53.28          \\ \hline
			\end{tabular}
		}
	\end{center}
	\label{tab:network_tranfer}
\end{table}

\subsection{Transferability}
Transferability refers to the capability of adversarial examples generated from one model to successfully attack another model, which is a way to achieve black-box attacks.
To evaluate the transferability of our attack method, we carry on the transferable experiments for hashing models with different backbone or different hash bits, which is summarized in Table \ref{tab:network_tranfer}.
We observe that the adversarial perturbations generated from one hash bit can achieve much similar t-MAP to another hash bit based on the same architecture of hashing model.
Besides, our method equips with good transferability from one DNN to another DNN while DHTA fails to transfer cross networks.
For example, when we adopt adversarial examples generated by ProS-GAN with H-AlexNet* to attack H-VGG11, an 8.84\% targeted performance increases for the \textit{Original} t-MAP, but the result of DHTA only changes by 0.47\%.

\subsection{Universality on different hashing methods}
We argue that our proposed scheme is applicable to most existing popular deep hashing methods. To evaluate this point, we compare with targeted attacks (P2P and DHTA) on other hashing methods, including DPH \cite{yang2018adversarial}, and HashNet \cite{cao2017hashnet}.
The results are reported in Table \ref{tab:hash_method}.
As shown in the table, even if tested on different deep hashing models, our targeted attack method is still effective and much better than the state-of-the-art DHTA in all cases, which further demonstrates the effectiveness of the proposed targeted hashing attack method.
For example, the t-MAP value of our ProS-GAN is higher than DHTA by 3.12\% on the HashNet with 32 bits code length.

\begin{table}[!t]
	\caption{t-MAP (\%) of targeted attack methods and MAP (\%) of benign samples for different hashing models on NUS-WIDE.}
	\begin{center}
		\setlength{\tabcolsep}{0.5mm}
		\resizebox{0.48\textwidth}{!}{
			\begin{tabular}{lc|cccc|cccc}
				\hline
				\multirow{2}{*}{Method} & \multirow{2}{*}{Metric} & \multicolumn{4}{c|}{DPH} & \multicolumn{4}{c}{HashNet}              \\ \cline{3-10}
				&                 & 12 bits & 24 bits & 32 bits & 48 bits  & 12 bits & 24 bits & 32 bits & 48 bits \\ \hline
				Original         & t-MAP    & 54.31 & 54.56 & 54.58 & 54.59 & 54.42 & 54.99 & 55.40 & 55.31  \\
				Noise            & t-MAP    & 53.12 & 53.40 & 53.42 & 53.45 & 53.51 & 53.99 & 54.09 & 54.10  \\
				P2P   & t-MAP & 69.66 & 70.79 & 71.00 & 71.38 & 65.16 & 69.28 & 71.72 & 73.17 \\
				DHTA  & t-MAP & 72.75 & 73.86 & 74.29 & 74.07 & 66.23 & 71.25 & 73.83 & 76.29 \\
				ProS-GAN       & t-MAP    & \textbf{75.57} & \textbf{76.87} & \textbf{78.62} & \textbf{77.62} & \textbf{71.29} & \textbf{75.06} & \textbf{76.95} & \textbf{78.48} \\ \hline
				Anchor code    & t-MAP    & 74.25 & 75.72 & 75.94 & 75.97 & 70.41 & 75.68 & 77.92 & 79.56  \\
				Prototype code  & t-MAP    & 78.01 & 79.83 & 79.85 & 78.79 & 73.99 & 78.58 & 80.67 & 81.67 \\ \hline
				Original         & MAP      & 70.09 & 71.27 & 71.23 & 71.50 & 66.40 & 70.69 & 72.62 & 73.78 \\ \hline
			\end{tabular}
		}
	\end{center}
	\label{tab:hash_method}
\end{table}

\section{Conclusion}
In this paper, we proposed a prototype-supervised adversarial network (ProS-GAN) for flexible targeted hashing attack, including a PrototypeNet, a generator and a discriminator. Specifically, we defined a category-level PrototypeNet to generate the semantic representation and to learn the prototype code as the representative of the target label for supervising the adversarial example generation. Moreover, the designed generator incorporated the decoded semantic representation into the original image to construct the adversarial example. Benefiting from the adversarial learning between the generator and the discriminator, the adversarial example could keep the visually realistic property and hold stronger attack performance. Extensive experiments showed that our ProS-GAN could achieve efficient and superior attack performance with higher transferability than the state-of-the-art targeted attack methods of deep hashing.

\section{Acknowledgments}
This work was supported in part by the National Natural Science Foundation of China (Grants Nos. 62002085, 62076213), the Guangdong
Basic and Applied Basic Research Foundation (Grants Nos. 2019A1515110475, 2019Bl515120055), and also supported by the university development fund of the Chinese University of Hong Kong, Shenzhen under grant No. 01001810, and the special project fund of Shenzhen Research Institute of Big Data under grant No. T00120210003.

\appendix
\section{Related Work: Optimization-based Attacks \& Generation-based Attacks}
In image classification, an adversarial example is usually a carefully modified image, which is intentionally perturbed by adding visually imperceptible perturbations to the original image but can confuse deep model to misclassify it. Since Szegedy \textit{et al.} \cite{szegedy2013intriguing} discovered the properties of adversarial examples, various adversarial attack methods in image classification have been proposed to fool a trained DNN. According to the information of target model exposed to the adversary, adversarial attacks can be categorized as white-box attacks (\eg, FGSM \cite{goodfellow2014explaining}, I-FGSM \cite{kurakin2016adversarial}, PGD \cite{madry2017towards} and C\&W \cite{carlini2017towards}) and black-box attacks (\eg SBA \cite{papernot2017practical} and ZOO \cite{chen2017zoo}).
For white-box attack, the adversary knows the whole network architecture and parameters so that it can design the adversarial perturbations by calculating the gradient of the loss \textit{w.r.t.} inputs.
For example, FGSM aims to increase the loss of the target model along the gradient direction once.
I-FGSM updates the perturbations multiple times with small step size and reaches better attack performance, which is an iterative variant of FGSM in essence.
For black-box attack, only the input and output are available to the adversary, thus it is difficult to get the gradient directly.
One solution is that we can use the transferability of adversarial examples to achieve black-box attacks. For instance, SBA adopts adversarial examples generated by a substitute model to attack the target model. Another way directly approximates the gradient base on the input data and output scores, such as ZOO.
Although black-box attack is difficult and its success rates are inferior to white-box attack, it is more general and practical.

The above attack methods are optimization-based, which regard the generation of adversarial examples as an optimization problem and use optimizers (\textit{e.g.}, box-constrained L-BFGS \cite{szegedy2013intriguing}) or gradient-based methods to solve it.
Optimization-based methods are powerful but quite slow because they need to access the target model iteratively for satisfactory attack performance.
Recently, generation-based attack methods received much more attention due to their high-efficiency during test phase.
Generation-based attack methods learn a generative model which transforms the input images into the corresponding adversarial samples.
Once the generative model trained, it do not need to access the target model again and can generate adversarial examples with one-forward pass.
Baluja \emph{et al.} \cite{baluja2017adversarial} firstly applied a generative model to take a original image as input and to generate its adversarial example.
Subsequently Xiao \emph{et al.} \cite{xiao2018generating} used GAN \cite{goodfellow2014generative} to produce adversarial examples with high perceptual quality.
Moreover, Mopuri \emph{et al.} \cite{mopuri2018nag} and Poursaeed \emph{et al.} \cite{poursaeed2018generative} proposed generative architectures to generate adversarial perturbations from any given random noise.
Finally, to achieve arbitrary target (category) attack, MAN \cite{han2019once} is designed a special generator which combines the features of the target label and the input image and outputs the targeted adversarial sample.

\begin{figure*}[!t]
	\begin{center}
		\includegraphics[width=0.98\linewidth]{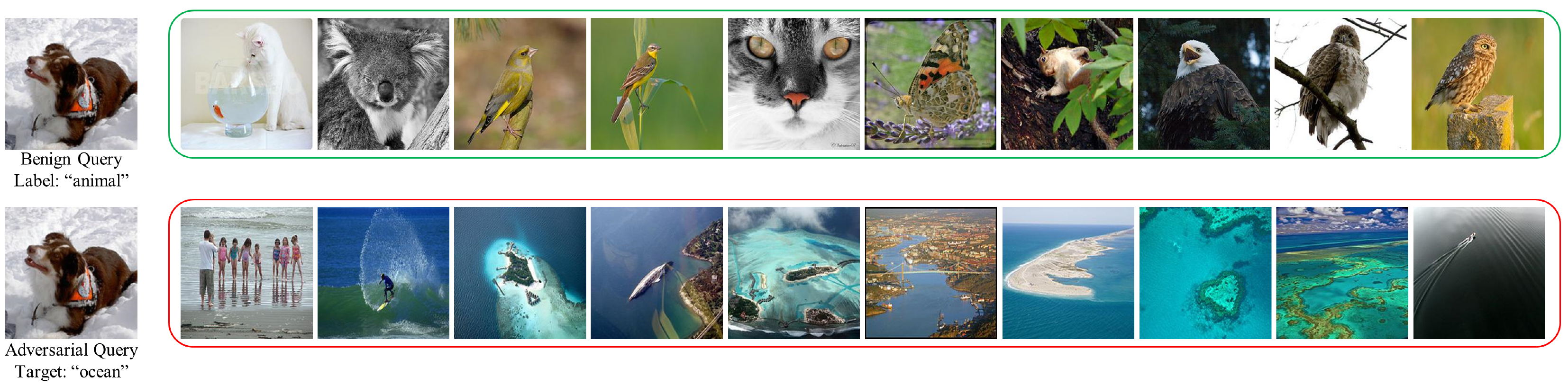}
	\end{center}
	\caption{An example to retrieve top-10 similarity samples on NUS-WIDE with the benign query and its adversarial query.}
	\label{fig:retrieval_10}
\end{figure*}

\section{Optimization}
The overall framework is actually a generative adversarial network, so the overall objective function can be written as a minimax optimization problem:
\begin{equation}
	\begin{aligned}
		(\theta_{p}, \theta_g) &= \mathop{\arg\min} \mathcal{L}_{pro}(\theta_{p}) + \mathcal{L}_{gen}(\theta_g) - \mathcal{L}_{dis}(\hat{\theta}_d) \\
		\theta_{d} &= \mathop{\arg\max} \mathcal{L}_{pro}(\hat{\theta}_{p}) + \mathcal{L}_{gen}(\hat{\theta}_g) - \mathcal{L}_{dis}(\theta_d) \\
		&\text { s.t. } B^{(p)} \in\{-1,1\}^{K \times M}.
	\end{aligned}
\end{equation}
Like all other generative adversarial networks, we optimize the entire network in an alternate way.
Firstly, when fixing $B$ and $L$, we optimize the $\mathcal{L}_{pro}$ over $\theta_p$.
Then, we optimize $\mathcal{L}_{gen}$ over $\theta_{g}$ by fixing the parameter $\theta_p$.
Finally, we optimize $\mathcal{L}_{dis}$ over $\theta_{d}$ by fixing the parameters $\theta_{p}$ and $\theta_{g}$.
The whole optimization process is outlined in algorithm \ref{alg:pros-gan} for detail.
Once the whole networks are trained in convergence, for any given target label and image, ProS-GAN can generate the corresponding adversarial example with a fast forward pass.

\renewcommand{\algorithmicrequire}{\textbf{Input:}}
\renewcommand{\algorithmicensure}{\textbf{Output:}}
\begin{algorithm}[t]
	\caption{Optimization procedure of ProS-GAN.}
	\label{alg:pros-gan}
	\begin{algorithmic}
		\REQUIRE
		Image dataset $O=\{(x_i, y_i)\}_{i=1}^N$, label set $L=\{y_i\}_{i=1}^M$, a pre-trained hashing model $F=sign(f_\theta(\cdot))$, and the hash code matrix $B$ for $O$ produced by $F$.
		\ENSURE
		Network parameters ($\theta_p, \theta_g, \theta_d$).
		\STATE \textbf{Initialize: } \\ Initialize parameters $\theta_p$, $\theta_g$, $\theta_d$, $\alpha_1$, $\alpha_2$, $\alpha_3$, $\alpha$, $\beta$ \\
		Learning rate $\eta$, batch size $n$
		\WHILE{not converge}
		\STATE Provide a batch of image set $\hat{O}$ and target labels $\hat{L}$
		\STATE Update $\theta_{p}$ by the gradient descent:
		\STATE \quad $\theta_{p} \gets \theta_{p} - \eta \Delta_{\theta_p} \frac{1}{n}(\mathcal{L}_{pro}+\mathcal{L}_{gen}-\mathcal{L}_{dis})$
		\STATE Update $\theta_{g}$ by the gradient descent:
		\STATE \quad $\theta_{g} \gets \theta_{g} - \eta \Delta_{\theta_g} \frac{1}{n}(\mathcal{L}_{pro}+\mathcal{L}_{gen}-\mathcal{L}_{dis})$
		\STATE Update $\theta_{d}$ by the gradient descent:
		\STATE \quad $\theta_{d} \gets \theta_{d} - \eta \Delta_{\theta_d} \frac{1}{n}(\mathcal{L}_{dis} - \mathcal{L}_{pro} - \mathcal{L}_{gen})$
		\ENDWHILE
	\end{algorithmic}
\end{algorithm}

\section{Discussion on Differences from the Related Works}
\textbf{Difference from P2P \cite{bai2020targeted} and DHTA \cite{bai2020targeted}.}
P2P and DHTA heuristically select a hash code from the set of hash codes of samples with the target label as objective code for targeted attack.
In contrast to P2P and DHTA, we design a prototype network (PrototypeNet) to learn the prototype code of the target label to supervise the generation of adversarial examples.
Because the PrototypeNet is designed to maximize the similarities of hash codes of samples with relevant labels and separability of those with irrelevant labels, the generated prototype code is the more representative and discriminative code of the hash codes of samples with the target label. In this way, they can be used as the target code to achieve more effective targeted attack performance.
In addition, compared to gradient-based hashing attack methods \cite{yang2018adversarial, bai2020targeted}, our proposed generation-based scheme is clearly faster to produce adversarial examples based on the optimization strategy, which is verified in the experiments section. Therefore, our method is intrinsically different from the existing algorithms, but more efficient and effective for targeted attack of deep hashing.

\textbf{Difference from MAN \cite{han2019once}.}
MAN designs a special generator to realize arbitrary-label targeted attacks on image classification model by combining input categories and images features. In contrast, our work is conceived for attacking deep hashing models. Due to the difference between classification and hashing, we design an effective PrototypeNet to learn semantic representation and prototype code. Furthermore, we upsample the semantic representation of the target label to the same dimension as the image, and then concatenation them together as the input of the encoder-decoder $G_{xt}$.
In addition, our proposed framework is essentially a generative adversarial network and we employ the adversarial learning between the generator and the discriminator to improve the visual quality of generated adversarial examples. Therefore, our work is totally different from MAN on problem definition, objective design, and framework construction.

\textbf{Difference from SSAH \cite{li2018self}.}
In terms of tasks, SSAH is formulated for cross-modal hashing, while ProS-GAN is used for adversarial attack. Particularly, SSAH employs the simple label network for generating semantic hash codes, which could be used to guide the image and text branches. As for differences between PrototypeNet and LabelNet in SSAH \cite{li2018self}, our PrototypeNet learns the prototype code from the hash codes produced by the attacked hashing model, while the LabelNet is a self-supervised network to generate semantically preserved hash codes. Moreover, they have different objectives on model design and feature learning purpose. Therefore, they are completely different.

\begin{figure*}[!t]
	\centering
	\subfigure[FLICKR-25K]{
		\begin{minipage}[t]{1.0\linewidth}
			\centering
			\includegraphics[width=1.0\linewidth]{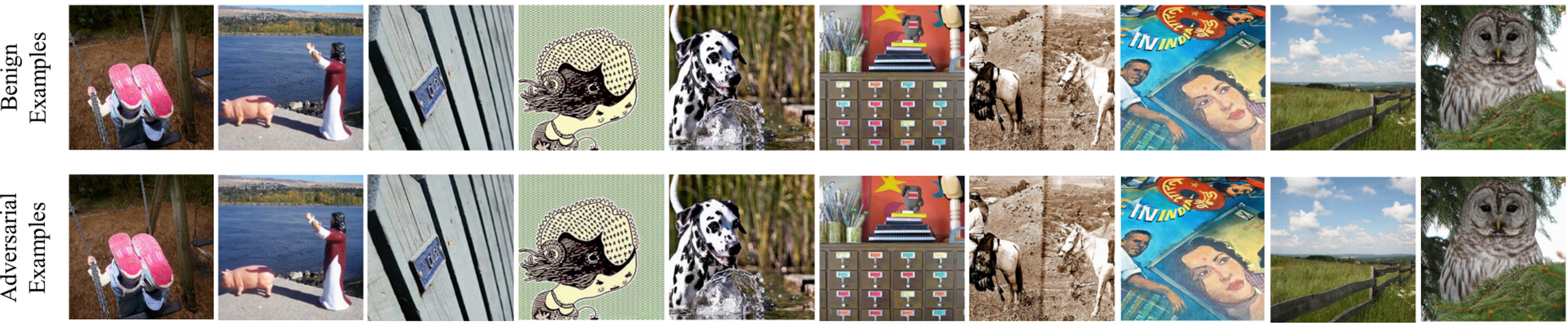}\\
		\end{minipage}%
	}%
	\quad
	\subfigure[NUS-WIDE]{
		\begin{minipage}[t]{1.0\linewidth}
			\centering
			\includegraphics[width=1.0\linewidth]{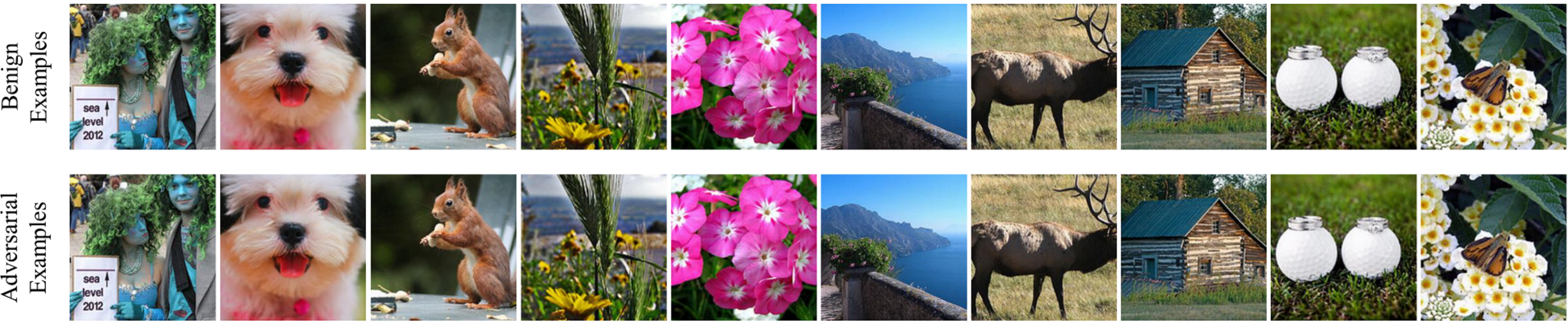}\\
		\end{minipage}%
	}%
	\quad
	\subfigure[MS-COCO]{
		\begin{minipage}[t]{1.0\linewidth}
			\centering
			\includegraphics[width=1.0\linewidth]{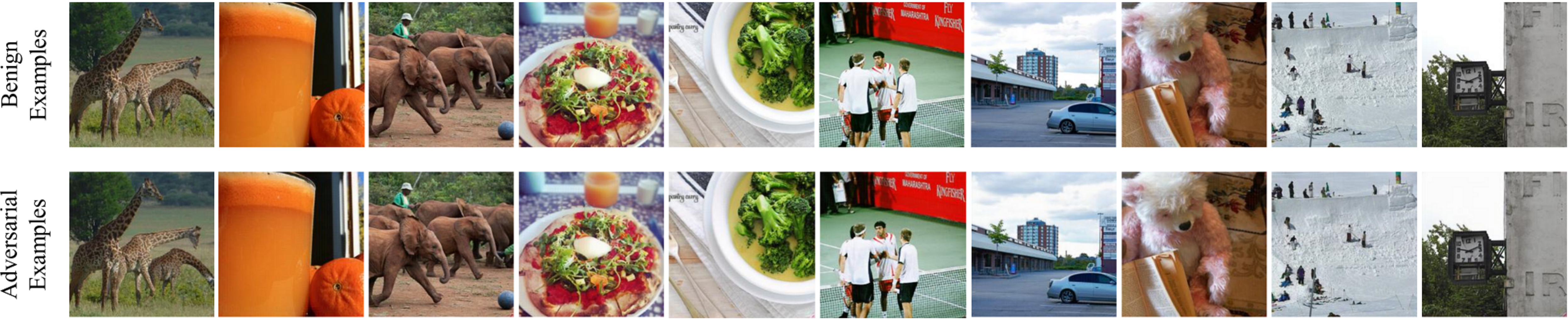}\\
		\end{minipage}%
	}%
	\centering
	\quad
	\caption{Visualization examples of generated adversarial samples.}
	\label{fig:visual}
\end{figure*}

\section{Visualization}
In this section, we provide some visual examples of the adversarial images generated by ProS-GAN on FLICKR-25K \cite{huiskes2008mir}, NUS-WIDE \cite{chua2009nus} and MS-COCO \cite{lin2014microsoft}. The comparison results are illustrated in Figure \ref{fig:visual}.
As we can see, the adversarial examples are almost the same as the benign examples (\ie original images).
An example of the retrieval results with a benign image and its adversarial example generated by our method is displayed in Figure \ref{fig:retrieval_10}.

\section{Transferability}
\textbf{Cross-hash bit transfer:}
In image hashing, the targeted adversarial perturbations generated from one hash bit can transfer to another hash bit based on the same architecture of hashing model, called cross-hash bit transfer \cite{yang2018adversarial}.
From Table \ref{tab:hash_bit_transfer}, we observe that the adversarial perturbations with different hash bits can achieve much similar t-MAP. We can see that the results of cross-hash bit transfer are superior to the state-of-the-art DHTA.

\begin{table}[!t]
	\caption{t-MAP (\%) of adversarial examples from one hash bit to another hash bits based on VGG11 backbone for NUS-WIDE.}
	\begin{center}
		\resizebox{0.45\textwidth}{!}{
			\setlength{\tabcolsep}{1.3mm}
			\begin{tabular}{c|c|cccc}
				\hline
				Method                & Code length & 12 bits & 24 bits & 32 bits & 48 bits \\ \hline
				\multirow{4}{*}{DHTA} & 12 bits     & 74.04   & 74.86   & 74.94   & 74.83   \\
				& 24 bits     & 73.62   & 75.60   & 75.71   & 75.69   \\
				& 32 bits     & 73.32   & 75.03   & 75.65   & 75.49   \\
				& 48 bits     & 72.52   & 74.19   & 74.69   & 75.63   \\ \hline
				\multirow{4}{*}{ours} & 12 bits     & 77.73   & 76.20   & 76.05   & 76.27   \\
				& 24 bits     & 76.60   & 78.21   & 78.39   & 78.25   \\
				& 32 bits     & 76.42   & 77.53   & 78.25   & 77.81   \\
				& 48 bits     & 75.19   & 76.06   & 76.71   & 78.75   \\ \hline
			\end{tabular}	
		}
	\end{center}
	\label{tab:hash_bit_transfer}
\end{table}

\textbf{Cross-network transfer:}
Cross-network transfer means that the adversarial perturbations computed from one DNN can attack another DNN successfully, which is also a black box attack.
In this section, we supplement the transfer results on FLICKR-25K and MS-COCO datasets, as summarized in Table \ref{tab:flickr_tranfer} and \ref{tab:coco_tranfer}.
We observe that the adversarial samples generated from one hash code length model has similar targeted attack performance to another code length model based on the same architecture, which is {cross-hash bit transfer} \cite{yang2018adversarial}.
For example, applying the adversarial examples generated by ProS-GAN from DH-AlexNet to attack DH-AlexNet* on MS-COCO (Table \ref{tab:coco_tranfer}) can achieve the similar t-MAP result (67.67\%) for 66.26\% of DH-AlexNet.
In most cases, the cross-hash bit transfer results of ProS-GAN are better than DHTA.
In addition, it is known that the advesarial examples computed from one backbone network can attack another network, called \textit{cross-network transfer} \cite{yang2018adversarial}.
Our ProS-GAN also has better cross-network transfer than DHTA.
For example, in Table \ref{tab:flickr_tranfer}, when we adopt the adversarial examples generated from DH-VGG11 to attack DH-ResNet18*, the t-MAP is 82.25\%, which is higher than 72.26\% of DHTA.
From these results, we conclude that the adversarial samples generated by our method have better transferability.

\begin{table*}[!t]
	\caption{Transfer t-MAP (\%) on FLICKR-25K dataset. DH-AlexNet, DH-VGG11 and DH-ResNet18 denote 12 bits DPSH models based on AlexNet \cite{krizhevsky2012imagenet}, VGG11 \cite{simonyan2014very} and ResNet18 \cite{he2016deep}, respectively, and "*" denotes their 32 bits variants.}
	\begin{center}
		\resizebox{1.0\textwidth}{!}{
			\setlength{\tabcolsep}{1.0mm}
			\begin{tabular}{ll|cccccc}
				\hline
				Method                    & Attacked model & DH-AlexNet & DH-AlexNet* & DH-VGG11 & DH-VGG11* & DH-ResNet18 & DH-ResNet18* \\ \hline
				\multirow{6}{*}{DHTA}     & DH-AlexNet      & 82.26     & 82.50      & 67.63   & 67.88    & 67.62      & 67.37       \\
				& DH-AlexNet*     & 81.89     & 84.08      & 67.68   & 68.18    & 67.52      & 67.56       \\
				& DH-VGG11        & 64.73     & 64.40      & 86.27   & 87.35    & 72.32      & 72.26       \\
				& DH-VGG11*       & 65.13     & 64.82      & 86.29   & 88.35    & 73.10      & 73.61       \\
				& DH-ResNet18     & 63.50     & 63.33      & 66.19   & 65.81    & 85.48      & 86.55       \\
				& DH-ResNet18*    & 63.42     & 63.27      & 65.92   & 65.61    & 85.38      & 87.80       \\ \hline
				\multirow{6}{*}{ProS-GAN} & DH-AlexNet      & 84.89     & 84.52      & 78.72   & 79.09    & 78.45      & 78.06       \\
				& DH-AlexNet*     & 80.95     & 81.99      & 77.44   & 77.48    & 76.72      & 77.43       \\
				& DH-VGG11        & 73.67     & 73.07      & 89.05   & 88.14    & 81.90      & 82.25       \\
				& DH-VGG11*       & 73.34     & 72.77      & 89.00   & 91.10    & 79.32      & 79.92       \\
				& DH-ResNet18     & 73.50     & 72.45      & 75.90   & 75.22    & 87.95      & 87.65       \\
				& DH-ResNet18*    & 72.42     & 72.95      & 76.18   & 75.97    & 86.25      & 88.19       \\ \hline
				\multicolumn{2}{l|}{Original}              & 62.83     & 62.61      & 63.58   & 63.49    & 63.23      & 63.20       \\ \hline
			\end{tabular}
		}
	\end{center}
	\label{tab:flickr_tranfer}
\end{table*}

\begin{table*}[!t]
	\caption{Transfer t-MAP (\%) on COCO dataset. DH-AlexNet, DH-VGG11 and DH-ResNet18 denote 12 bits DPSH models based on AlexNet \cite{krizhevsky2012imagenet}, VGG11 \cite{simonyan2014very} and ResNet18 \cite{he2016deep}, respectively, and "*" denotes their 32 bits variants.}
	\begin{center}
		\resizebox{1.0\textwidth}{!}{
			\setlength{\tabcolsep}{1.0mm}
			\begin{tabular}{ll|cccccc}
				\hline
				Method                    & Attacked model & DH-AlexNet & DH-AlexNet* & DH-VGG11 & DH-VGG11* & DH-ResNet18 & DH-ResNet18* \\ \hline
				\multirow{6}{*}{DHTA}     & DH-AlexNet      & 57.05     & 58.39      & 45.23   & 45.83    & 45.03      & 45.28       \\
				& DH-AlexNet*     & 55.88     & 58.35      & 45.15   & 45.86    & 44.94      & 45.24       \\
				& DH-VGG11        & 44.49     & 45.16      & 59.85   & 61.78    & 51.61      & 51.56       \\
				& DH-VGG11*       & 44.35     & 44.95      & 59.12   & 63.22    & 50.38      & 51.18       \\
				& DH-ResNet18     & 42.98     & 43.77      & 44.56   & 44.77    & 61.88      & 64.44       \\
				& DH-ResNet18*    & 42.95     & 43.73      & 44.13   & 44.54    & 62.12      & 65.42       \\ \hline
				\multirow{6}{*}{ProS-GAN} & DH-AlexNet      & 66.26     & 67.67      & 53.79   & 55.57    & 54.22      & 55.22       \\
				& DH-AlexNet*     & 66.49     & 69.41      & 53.73   & 55.45    & 54.56      & 55.14       \\
				& DH-VGG11        & 49.75     & 51.00      & 66.22   & 65.35    & 56.14      & 54.49       \\
				& DH-VGG11*       & 50.14     & 51.21      & 67.08   & 71.65    & 58.23      & 57.48       \\
				& DH-ResNet18     & 49.30     & 49.67      & 49.47   & 51.23    & 70.27      & 72.07       \\
				& DH-ResNet18*    & 48.87     & 50.43      & 49.36   & 51.60    & 68.75      & 72.95       \\ \hline
				\multicolumn{2}{l|}{Original}              & 42.41     & 43.24      & 42.33   & 42.67    & 42.40      & 42.85       \\ \hline
			\end{tabular}
		}
	\end{center}
	\label{tab:coco_tranfer}
\end{table*}

{\small
	\bibliographystyle{ieee_fullname}
	\bibliography{egbib}
}

\end{document}